\documentclass[10pt,twocolumn,letterpaper]{article}

\usepackage{iccv}
\usepackage{times}
\usepackage{epsfig}
\usepackage{graphicx}
\usepackage{amsmath}
\usepackage{amssymb}
\usepackage{algorithm}
\usepackage{amsthm}
\newtheorem{lemma}{Lemma}
\newtheorem{theorem}{Theorem}
\usepackage{multirow}

\usepackage[pagebackref=true,breaklinks=true,letterpaper=true,colorlinks,bookmarks=false]{hyperref}

\iccvfinalcopy 

\def\iccvPaperID{****} 

\begin{document}

\title{Do not Omit Local Minimizer: a Complete Solution for Pose Estimation from 3D Correspondences }

\author{Lipu Zhou$^1$, Shengze Wang$^1$, Jiamin Ye$^2$, Michael Kaess$^1$\\
	$^1$Carnegie Mellon University, $^2$Chinese Academy of Sciences\\
	{\tt\small $^1$zhoulipu@outlook.com, $^1$\{shengzew,kaess\}@andrew.cmu.edu, $^2$yejiamin@iet.cn}}

\maketitle

\begin{abstract}
Estimating pose from given 3D correspondences, including point-to-point, point-to-line and point-to-plane correspondences, is a fundamental task in computer vision with many applications. We present a complete solution for this task, including a solution for the minimal problem and the least-squares problem of this task. Previous works mainly focused on finding the global minimizer to address the least-squares problem. However, existing works that show the ability to achieve global minimizer are still unsuitable for  real-time applications. Furthermore, as one of contributions of this paper, we prove that there exist ambiguous configurations for any number of lines and planes. These configurations have several solutions in theory, which makes the  correct solution may come from a local minimizer.  Our algorithm is efficient and able to reveal local minimizers.  We employ the Cayley-Gibbs-Rodriguez (CGR) parameterization of the rotation  to derive a general rational cost for the three cases of 3D correspondences. The main contribution of this paper is to solve the resulting equation system of the minimal problem and the  first-order optimality conditions of the least-squares problem, both of which are of complicated rational forms. The central idea of our algorithm is to introduce intermediate unknowns to simplify the problem. Extensive experimental results show that our algorithm significantly outperforms previous algorithms when the number of correspondences is small. Besides,  when the global minimizer is the solution, our algorithm achieves the same accuracy as previous algorithms that have guaranteed global optimality, but our algorithm is applicable to real-time applications.
\end{abstract}

\section{Introduction}

Estimating the pose from 3D correspondences, \ie point-to-point, point-to-line and point-to-plane correspondences, is known as the 3D registration problem in the literature \cite{olsson2006registration,olsson2008solving,olsson2009branch,wientapper2016unifying,briales2017convex,wientapper2018universal}.  It is one of the fundamental problems in computer vision with a wide range of applications, such as simultaneous localization and mapping (SLAM)~\cite{zhang2015visual,zhang2014loam,newcombe2011kinectfusion,proencca2018probabilistic,taguchi2013point}, extrinsic calibration~\cite{zhang2004extrinsic,unnikrishnan2005fast,naroditsky2011automatic,gomez2015extrinsic,zhou2018automatic} and iterative closes point (ICP) framework~\cite{besl1992method}. Besides,  some camera pose estimation problems, such as the perspective-n-point (PnP) problem~\cite{hesch2011direct,kneip2014upnp} and the perspective-n-line (PnL) problem~\cite{xu2017pose,pvribyl2017absolute},  can be transformed to a 3D registration problem \cite{wientapper2018universal}. However, the research on this problem is not as thorough as other pose estimation problems. 

Previous works mainly focus on solving the least-squares problem. Although large progress has been made, achieving the optimal solution and real-time performance is still a challenge.  Some algorithms \cite{olsson2009branch,briales2017convex} are capable of  finding the global minimum, however, their running time makes them not suitable for real-time applications. Recently, Wientapper \cite{wientapper2018universal} provided an efficient algorithm, but their algorithm can not achieve the global optimality in theory, and has the risk of no solution. Furthermore, previous works generally assume that the correct solution  is the global minimizer. However, we find that the correct solution may come from a local minimizer in certain configurations.

Minimal solution is important to eliminate outliers in the RANSAC framework \cite{fischler1981random}. As the pose has 6 degrees of freedom (DoF), any combinations of  3D correspondences providing 6 independent constraints form a minimal configuration.  Previous algorithms are proposed to solve these minimal configurations case by case  \cite{chen1990pose,ramalingam2010p2pi,ramalingam2013theory}. The contributions of this paper are as follows:

First, we prove that there exist ambiguous configurations for any number of plane and line correspondences, which leads to multiple solutions. When a configuration approximates ambiguity, the correct solution of a problem may come from a local minimum. Therefore, previous works \cite{olsson2006registration,olsson2009branch,briales2017convex} that only  compute the global minimizer will fail in this case. Revealing local minimizers is essential for an algorithm to handle all the configurations.

Second, we present an efficient and accurate solution for the least-squares 3D registration problem.  We use the  CGR parametrization to represent the rotation, which generates a rational cost function. We derive its first-order optimality conditions. They form a  high order polynomial system, and are hard to solve. Four intermediate unknowns are introduced to relax the original problem, which results in  much simpler first-order optimality conditions.   Gr\"obner basis method \cite{cox2006using} is applied to solve this equation system. Then we refine the solution by the Newton-Raphson method. 

Third, we present an unified  solution for the potential minimal configurations. Previous algorithms are proposed to solve the minimal problems case by case \cite{chen1990pose,ramalingam2010p2pi,ramalingam2013theory}. We also use the CGR parametrization to represent the rotation. which generates a third order equation system for the minimal configurations. Three intermediate unknowns are introduced to simplify the equation system. We introduce a novel hidden variable method \cite{cox2006using} to solve the resulting second order equation system for the rotation matrix. Then the translation can be computed from a linear system.

We evaluate our algorithm with synthetic and real data. Extensive experimental results show that recovering the local minimal is essential, especially when the number of correspondences $N$ is small. For a small $N$, our algorithm significantly outperforms previous works \cite{olsson2006registration,olsson2009branch,briales2017convex,wientapper2018universal}.  
Besides, experimental results verify that  our algorithm can converge to the global minimizer with  the same accuracy as  the previous work with guaranteed globally optimality \cite{olsson2009branch,briales2017convex}, however, our algorithm is much faster. 



\section{Related Works}
This paper focuses on the pose estimation from  point-to-point, point-to-line and point-to-plane correspondences correspondences. Pose estimation from point-to-point correspondences has been solved in early works \cite{arun1987least,horn1987closed,horn1988closed}. There exist closed-form solutions for this problem. However, estimating pose from point-to-line and point-to-plane correspondences is more complicated. These 3D correspondences actually yield similar  distance function, which is employed by previous works to construct a general cost function \cite{olsson2006registration,olsson2008solving,olsson2009branch,wientapper2016unifying,briales2017convex,wientapper2018universal}. The difference of various cost functions lies in the parameterization of the rotation matrix. The raw rotation matrix is adopted in \cite{olsson2008solving,briales2017convex}, and the quaternion \cite{horn1987closed} is used in \cite{olsson2006registration,olsson2009branch,wientapper2016unifying,wientapper2018universal}.  Previous works mainly focus on finding the global minimizer of the resulting cost function. In \cite{olsson2009branch}, they proposed a provably optimal algorithm. They employed convex underestimators with branch-and-bound methods to iteratively compute the global minimum of the cost function. Although this algorithm can guarantee global optimality, it is very time-consuming. Olsson \etal \cite{olsson2008solving} reduced the computational complexity by  applying the Lagrangian dual relaxation to approximate the cost function. Their experimental results showed that a single convex semidefinite program can well approximate  the original problem. Recently, an improved result is obtained by a strengthened Lagrangian dual relaxation \cite{briales2017convex}. Although without theoretical guarantees, their experimental results show that this algorithm can achieve the global minimum.  Although these algorithms \cite{olsson2008solving,briales2017convex} have made progress in efforts to reduce the computational time, they are still not suitable to demanding real-time applications. Recently, Wientapper \cite{wientapper2018universal} provided an efficient algorithm based on Gr\"obner basis polynomial solver. They  introduced the first order derivatives of the norm-one constraint of the quaternion  
into the first order optimality conditions of the cost function as \cite{kneip2014upnp}, rather than apply the Lagrangian formulation for the norm-one constraint of the quaternion. 
This results in an efficient solution, however, their result is not optimal in theory.  Furthermore, as the number of equations is greater than the number of unknowns, the equation system may  have no solution.  They introduced 4 pre-rotations to solve the problem 4 times to handle this problem. The pre-rotation is  to make the no-solution problem happen less likely, however, it can not fully solve this problem in theory.

Minimal solution is essential in the RANSAC framework \cite{fischler1981random} for robust pose estimation. Any combination of the 3D correspondences resulting in 6 independent constraints forms a minimal configuration. In the literature, specific algorithms have been proposed to solve certain configurations.    In \cite{87340}, a minimal solution for three line-to-plane correspondences were proposed. In \cite{vasconcelos2012minimal,zhou2014new}, they studied the minimal solution of a more specific line-to-plane configuration, \ie, lines are all on a plane. As one line-to-plane correspondence can be treated as two point-to-plane correspondences, these problems can be formulated as a point-to-plane registration problem. Naroditsky \etal \cite{naroditsky2011automatic} presented a solution to 6 point-to-plane correspondences.  Ramalingam \etal \cite{ramalingam2013theory} solved all the point-to-plane minimal configurations. They designed  specific intermediate coordinate systems to simplify  each minimal configuration.  This method is not convenient in the application, as it needs to implement a bunch of algorithms. 

3D registration without correspondence is an related but more complicated problem.  Several works \cite{li20073d,papazov2011stochastic,yang2013go,zhou2016fast} presented globally optimal solution for  3D point registration without correspondence.   The ICP framework \cite{besl1992method} gives a general way to find  correspondences and pose at the same time. Although this framework was originally designed for points, other 3D models can also be introduced into this framework \cite{segal2009generalized,censi2008icp,serafin2015nicp}. The ICP framework iteratively finds the nearest elements as correspondences then calculates the pose until it converges. Efficiently and accurately estimating the pose from current nearest 3D correspondences is critical for the ICP framework. 


\section{Problem Formulation}\label{sec:Problem}


In this paper, we use italic, boldfaced lowercase and boldfaced uppercase letters to represent scalars, vectors and matrices, respectively. The paper focuses on the problem of pose estimation from point-to-plane, point-to-line and point-to-point correspondences.
We first consider the point-to-plane correspondence. Suppose we have a point ${{\mathbf{x}}_{l}}$ in one coordinate system and  the corresponding plane in another coordinate system represented by the plane's norm-one normal $\mathbf{n}$ and a point ${{\mathbf{y}}^{\pi }}$ lying on it. Then the scalar residual between a point and a plane can be written as
\begin{equation} \label{equ:1}
{{r}_{\pi }}={{\mathbf{n}}^{T}}\left( \mathbf{R}{{\mathbf{x}}_{\pi }}+\mathbf{t}-{{\mathbf{y}}_{\pi }} \right)
\end{equation}
For the point-to-line correspondence, we can calculate the 3-dimensional residual vector between a point ${{\mathbf{x}}_{l}}$ and the corresponding line represented by the norm-one direction $\mathbf{d}$ and a point ${{\mathbf{y}}_{l}}$  on it as 
\begin{equation} \label{equ:2}
{{\mathbf{r}}_{l}}=\left( {{\mathbf{I}}_{3}}-\mathbf{d}{{\mathbf{d}}^{T}} \right)\left( \mathbf{R}{{\mathbf{x}}_{l}}+\mathbf{t}-{{\mathbf{y}}_{l}} \right)
\end{equation}
where ${{\mathbf{I}}_{3}}$ is the identify matrix.
Lastly, a point-to-point correspondence yields a $3$-dimensional residual vector
\begin{equation} \label{equ:3}
{{\mathbf{r}}_{p}}=\mathbf{R}{{\mathbf{x}}_{p}}+\mathbf{t}-{{\mathbf{y}}_{p}}
\end{equation} 

$\mathbf{R}$ and $\mathbf{t}$ have 6 DoF. Our problem is to calculate   $\mathbf{R}$ and $\mathbf{t}$ from any configuration of ${{n}_{\pi }}$  point-to-plane, ${{n}_{l}}$ point-to-line, and ${{n}_{p}}$ point-to-point correspondences that has at least 6  effective constraints. Let use define  $N$ as the  number of effective correspondences. When  $N > 6$, this is a least squares problem described in section \ref{sec:ls}. When $N=6$, this forms a minimal problem solved in section \ref{sec:min}. 

\section{Least Squares Solution} \label{sec:ls}
Using the notation in (\ref{equ:1}), (\ref{equ:2}) and (\ref{equ:3}), we define our cost function as follows
\begin{equation} \label{equ:4}
\begin{array}{l}
{C_{\pi lp}}\left({\bf{R}},{\bf{t}}\right) = \sum\limits_{i = 1}^{{n_\pi }} {r_{{\pi _i}}^2}  + \sum\limits_{i = 1}^{{n_l}} {{\bf{r}}_{{l_i}}^T{{\bf{r}}_{{l_i}}}}  + \sum\limits_{i = 1}^{{n_p}} {{\bf{r}}_{{p_i}}^T{{\bf{r}}_{{p_i}}}} ,\\
s.t.\;{\bf{R}}{{\bf{R}}^T} = {{\bf{I}}_3},\det \left( {\bf{R}} \right) = 1,
\end{array}
\end{equation}
where $\det \left( \mathbf{R} \right)$ represents the determinant of $\mathbf{R}$.


\subsection{Ambiguous Configurations}
In previous works, the global minimizer is treated as the optimal solution. It is well known that there is an unique solution for at least three point-to-point correspondences, except for the degenerate configuration, \ie all the points are collinear. However, for plane and line, there exist configurations which have several solutions. We call them as ambiguous configurations.   The ambiguous configuration differs from the degenerate configuration. The ambiguous configuration has several solutions, but the degenerate configuration has infinite solutions. Formally, we have the following lemma for point-to-line correspondences. \textbf{Proof is in the supplementary material.}
\begin{lemma} \label{lemma:line}
	For any $n_l$ points and any $\left\{ {{{\bf{R}}_1},{{\bf{t}}_1}} \right\}$ and $\left\{ {{{\bf{R}}_2},{{\bf{t}}_2}} \right\}$, there exist  $n_l$ lines to make $\left\{ {{{\bf{R}}_1},{{\bf{t}}_1}} \right\}$ and $\left\{ {{{\bf{R}}_2},{{\bf{t}}_2}} \right\}$ are exact solutions for the $n_l$ point-to-line correspondences.
\end{lemma}
Similarly, we have lemma \ref{lemma:plane} for point-to-plane correspondences.
\begin{lemma} \label{lemma:plane}
	For any $n_{\pi}$ points and any $\left\{ {{{\bf{R}}_1},{{\bf{t}}_1}} \right\}$, $\left\{ {{{\bf{R}}_2},{{\bf{t}}_2}} \right\}$ and $\left\{ {{{\bf{R}}_3},{{\bf{t}}_3}} \right\}$, there exist  $n_{\pi}$ planes to make $\left\{ {{{\bf{R}}_1},{{\bf{t}}_1}} \right\}$, $\left\{ {{{\bf{R}}_2},{{\bf{t}}_2}} \right\}$ and $\left\{ {{{\bf{R}}_3},{{\bf{t}}_3}} \right\}$ are exact solutions for the $n_{\pi}$ point-to-plane correspondences.
\end{lemma}
Finally, we have the following theorem for  $n_l$ point-to-line and $n_{\pi}$ point-to-plane correspondences.
\begin{theorem} \label{theorm:1}
	For any $n_l$ points on lines and $n_{\pi}$ points on planes and any $\left\{ {{{\bf{R}}_1},{{\bf{t}}_1}} \right\}$ and $\left\{ {{{\bf{R}}_2},{{\bf{t}}_2}} \right\}$, there exist  $n_l$ lines and $n_{\pi}$ planes  to make $\left\{ {{{\bf{R}}_1},{{\bf{t}}_1}} \right\}$ and $\left\{ {{{\bf{R}}_2},{{\bf{t}}_2}} \right\}$ are exact solutions for the $n_l$ point-to-line and $n_{\pi}$ point-to-plane correspondences.
\end{theorem}


When measurements approximate an ambiguous configuration, a prior is required to identify the correct solution. The prior is generally available in real applications. For instance, we generally have a rough estimation of the pose between two sensors in the extrinsic calibration problem \cite{zhang2004extrinsic,unnikrishnan2005fast,naroditsky2011automatic,gomez2015extrinsic,zhou2018automatic}. For the pose estimation in SLAM \cite{zhang2015visual,zhang2014loam,newcombe2011kinectfusion,proencca2018probabilistic,taguchi2013point}, current pose should be consistent with the previous motion trajectory.

Previous works  \cite{olsson2006registration,olsson2008solving,olsson2009branch,briales2017convex} using convex approximation can only find the global minimizer. However, the global minimizer may not be the correct solution of a problem if measurements approximate an ambiguous configuration. Therefore, we can not simply omit local minimizers. 

\subsection{Rotation Parameterization}\label{sec:Algorithm}
Solving for the rotation matrix is the crux for  pose estimation. 
Previous works \cite{olsson2006registration,olsson2009branch,briales2017convex,wientapper2018universal} adopt non-minimal representations for $\bf{R}$, which results in additional quadratic constraints in the minimization problem. This paper adopts the CGR parametrization \cite{hesch2011direct,mirzaei2011optimal} which gives a minimal representation for the rotation matrix, removing the quadratic constraints in (\ref{equ:4}) . The CGR parametrization expresses a rotation matrix as 
\begin{equation} \label{equ:5}
{\mathbf{R}=\frac{{\mathbf{\bar{R}}}}{1+{{\mathbf{s}}^{T}}\mathbf{s}},\ \mathbf{\bar{R}}=\left( \left( 1-{{\mathbf{s}}^{T}}\mathbf{s} \right){{\mathbf{I}}_{3}}+2{{\left[ \mathbf{s} \right]}_{\times }}+2\mathbf{s}{{\mathbf{s}}^{T}} \right)}
\end{equation}
where $\mathbf{s}=\left[ {{s}_{1}};{{s}_{2}};{{s}_{3}} \right]$ is a 3-dimensional vector, and ${ \left[ {\bf{s}} \right]_ \times }$ is the
skew-symmetric matrix of $\bf{s}$.

\subsection{Rational Formulation of Residual} 
We will show that the residuals from point-to-point, point-to-line and point-to-plane correspondences have a general form
\begin{equation} \label{equ:6}
{{r}_{g}}={{\mathbf{a}}^{T}}\mathbf{Rb}+{{\mathbf{a}}^{T}}\mathbf{t}+c
\end{equation}
where $\mathbf{a}$ and $\mathbf{b}$ are two $3$-dimensional vectors and $c$ is a scalar.
We first consider the point-to-plane residual ${{r}_{\pi }}$ in (\ref{equ:1}). It is obvious that $\mathbf{a}=\mathbf{n},\ \mathbf{b}={{\mathbf{y}}_{\pi }}$ and $c=-{{\mathbf{n}}^{T}}{{\mathbf{y}}_{\pi }}$.
Let us then consider the point-to-line residual. We define ${{\mathbf{r}}_{l}}=\left[ r_{l}^{1},r_{l}^{2},r_{l}^{3} \right]^T$ and ${{\mathbf{A}}_{l}}=\left( {{\mathbf{I}}_{3}}-\mathbf{d}{{\mathbf{d}}^{T}} \right)=\left[ \mathbf{a}_{l}^{1};\mathbf{a}_{l}^{2};\mathbf{a}_{l}^{3} \right]$, where $\mathbf{a}_{l}^{1},\mathbf{a}_{l}^{2}$ and $\mathbf{a}_{l}^{3}$ are the three rows of ${{\mathbf{A}}_{l}}$. Using this notation, equation (\ref{equ:2}) can be written as
\begin{equation} \label{equ:7}
r_{l}^{i}=\mathbf{a}_{l}^{i}\left( \mathbf{R}{{\mathbf{x}}_{l}}+\mathbf{t}-{{\mathbf{y}}_{l}} \right),\ i=1,2,3
\end{equation}
It is easy to find that $\mathbf{a}={{\left( \mathbf{a}_{l}^{i} \right)}^{T}},\ \mathbf{b}={{\mathbf{x}}_{l}}$ and $c=-\mathbf{a}_{l}^{i}{{\mathbf{y}}_{\pi }}$.
Lastly, we consider the point-to-point correspondence. Similar to the point-to-line correspondence, we define ${{\mathbf{r}}_{p}}={{\left[ r_{p}^{1},r_{p}^{2},r_{p}^{3} \right]}^{T}}$ and ${{\mathbf{I}}_{3}}=\left[ \mathbf{e}_{l}^{1};\mathbf{e}_{l}^{2};\mathbf{e}_{l}^{3} \right]$. Substituting this notation into (\ref{equ:3}), it is easy to find
\begin{equation} \label{equ:8}
r_{p}^{i}=\mathbf{e}_{p}^{i}\left( \mathbf{R}{{\mathbf{x}}_{p}}+\mathbf{t}-{{\mathbf{y}}_{p}} \right),\ i=1,2,3
\end{equation}
The similarity between (\ref{equ:8}) and (\ref{equ:7}) is obvious. Therefore, $r_{p}^{i}$ also has the general form (\ref{equ:6}).

Let $\mathbf{t}=\left[ {{t}_{1}};{{t}_{2}};{{t}_{3}} \right]$. Substituting (\ref{equ:5}) into the general residual ${{r}_{g}}$ in (\ref{equ:6}) and adding the tree terms together yields
\begin{equation} \label{equ:9}
{r_g} = \frac{{{{\bf{k}}^T}{\bf{v}}}}{{1 + {{\bf{s}}^T}{\bf{s}}}}
\end{equation}
where ${{\mathbf{k}}^{T}}\mathbf{v}$  is a third order polynomial with terms ${\bf{v}} = [s_1^2{t_1},s_1^2{t_2},s_1^2{t_3},s_1^2,{s_1}{s_2},{s_1}{s_3},{s_1},s_2^2{t_1},s_2^2{t_2},s_2^2{t_3}, s_2^2,{s_2}{s_3}, \\ {s_2},s_3^2{t_1},s_3^2{t_2},s_3^2{t_3},s_3^2,{s_3},{t_1},{t_2},{t_3},1{]^T}$.

\subsection{First-Order Optimality Conditions}

Now we consider the least-squares problem (\ref{equ:4}). Using (\ref{equ:9}), the squared residual can be represented as

\begin{equation} \label{equ:10}
r_g^2 = \frac{{q\left( {{\bf{s}},{\bf{t}}} \right)}}{{{{\left( {1 + {{\bf{s}}}^T{\bf{s}}} \right)}^2}}}
\end{equation}
where $q\left( \mathbf{s},\mathbf{t} \right)={{\mathbf{v}}^{T}}\mathbf{k}{{\mathbf{k}}^{T}}\mathbf{v}$ is a 6th order polynomial in $\mathbf{s}$ and $\mathbf{t}$. For the point-to-line distance $\mathbf{r}_{l_i}^{T}{{\mathbf{r}}_{l_i}}$ in (\ref{equ:4}), we have 
\begin{equation} \label{equ:11}
\mathbf{r}_{l_i}^{T}{{\mathbf{r}}_{l_i}}={{\left( r_{l_i}^{1} \right)}^{2}}+{{\left( r_{l_i}^{2} \right)}^{2}}+{{\left( r_{l_i}^{3} \right)}^{2}}
\end{equation}
As mentioned above, $r_{l_i}^{1},$ $r_{l_i}^{2}$ and $r_{l_i}^{3}$ all have the same form as ${{r}_{g}}$. Therefore,  the point-to-line distance $\mathbf{r}_{l_i}^{T}{{\mathbf{r}}_{l_i}}$ will have the same form as (\ref{equ:10}). Similarly,  
$\mathbf{r}_{p}^{T}{{\mathbf{r}}_{p}}$ will also yield the same form as (\ref{equ:10}). 
After the summation of squared residuals in (\ref{equ:4}), we know that ${{C}_{\pi lp}}$ would have the same rational form as (\ref{equ:10}). We write it as
\begin{equation} \label{equ:12}
{{C}_{\pi lp}}\left(\mathbf{s},\mathbf{t}\right)=\frac{{{{\bar{C}}}_{\pi lp}}\left( \mathbf{s},\mathbf{t} \right)}{{{\left( 1+{{\mathbf{s}}^{T}{\mathbf{s}}} \right)}^{2}}}
\end{equation}
where ${{\bar{C}}_{\pi lp}}$ is a 6th order polynomial function in $\mathbf{s}$ and $\mathbf{t}$.
To find the critical points of (\ref{equ:12}), we first calculate  its first order optimality conditions as follows: 
\begin{equation} \label{equ:13}
\begin{array}{l}
{g_{{s_i}}} = \frac{{\partial {C_{\pi lp}}}}{{\partial {s_i}}} = \frac{1}{{{{\left( {1 + {{\bf{s}}^T}{\bf{s}}} \right)}^3}}}\left( {\left( {1 + {{\bf{s}}^T}{\bf{s}}} \right)\frac{{\partial {{\bar C}_{\pi lp}}}}{{\partial {s_i}}} - 4{s_i}{{\bar C}_{\pi lp}}} \right) = 0\\
{g_{{t_i}}} = \frac{{\partial {C_{\pi lp}}}}{{\partial {t_i}}} = \frac{1}{{{{\left( {1 + {{\bf{s}}^T}{\bf{s}}} \right)}^2}}}\frac{{\partial {{\bar C}_{\pi lp}}}}{{\partial {t_i}}} = 0,\;i = 1,2,3
\end{array}
\end{equation}
Canceling the denominator of (\ref{equ:13}) yields
\begin{equation} \label{equ:14}
\begin{array}{l}
{{\bar g}_{{s_i}}} = \left( {1 + {{\bf{s}}^T}{\bf{s}}} \right)\frac{{\partial {{\bar C}_{\pi lp}}}}{{\partial {s_i}}} - 4{s_i}{{\bar C}_{\pi lp}} = 0\\
{{\bar g}_{{t_i}}} = \frac{{\partial {{\bar C}_{\pi lp}}}}{{\partial {t_i}}} = 0,\;i = 1,2,3
\end{array}
\end{equation}
$\frac{{\partial {{\bar C}_{\pi lp}}}}{{\partial {s_i}}}$ and $\frac{{\partial {{\bar C}_{\pi lp}}}}{{\partial {t_i}}}$ are of degree 5. Therefore, ${{\bar{g}}_{{{s}_{i}}}}$ and ${{\bar{g}}_{{{t}_{i}}}}$ are of degree 7 and 5, respectively. Although the Gr\"obner basis method gives a general way to solve the polynomial system, it is computationally demanding and numerically unstable to apply it to a high order polynomial system, as the experimental results in Fig. \ref{fig:vs_DLSSol}. The Newton-Raphson method provides an alternative way to find the roots of (\ref{equ:14}).  Denote an equation system as $\bar{G}\left({\bf{x}}\right)$. For the $k$th iteration, the Newton-Raphson method updates the solution as  
\begin{equation} \label{equ:15}
{{\bf{x}}_k}  = {{\bf{x}}_{k-1}}  - J_{\bar G}^{ - 1}\left( {{\bf{x}}_{k-1}} \right)\bar G\left( {{\bf{x}}_{k-1}}  \right)
\end{equation}

This iterative method requires an initial solution ${{\mathbf{s}}_{0}}$ and ${{\mathbf{t}}_{0}}$. In the next section, we will describe how to calculate an accurate ${{\mathbf{s}}_{0}}$ and ${{\mathbf{t}}_{0}}$.

\subsection{Initial Estimation from Relaxation}

The difficulty of solving (\ref{equ:14}) lies in the denominator of CGR parameterization (\ref{equ:5}). To solve this problem, we introduce the following intermediate variable
\begin{equation} \label{equ:16}
\rho  = \frac{1}{{\sqrt {1 + s_1^2 + s_2^2 + s_3^2} }}.
\end{equation}
Substituting (\ref{equ:16}) into (\ref{equ:9}) will transform (\ref{equ:9}) from a rational function to a polynomial function. Furthermore,  we define
\begin{equation} \label{equ:17}
\alpha =\rho {{s}_{1}},\ \beta =\rho {{s}_{2}},\ \gamma =\rho {{s}_{3}}.
\end{equation}
Using (\ref{equ:16}) and (\ref{equ:17}), the residual (\ref{equ:9}) can be expressed as 
\begin{equation} \label{equ:18}
{{r}_{g}}=\mathbf{a}_{g}^{T}\mathbf{u}+{{\mathbf{b}_{g}}^{T}}\mathbf{t}
\end{equation}
where $\mathbf{u}={{\left[ {{\alpha }^{2}},\alpha \beta ,\alpha \gamma ,\alpha \rho ,{{\beta }^{2}},\beta \gamma ,\beta \rho ,{{\gamma }^{2}},\gamma \rho ,{{\rho }^{2}},1 \right]}^{T}}$. This formulation turns the rational function  (\ref{equ:9}) into a polynomial function (\ref{equ:18}) that is easier to handle.  Furthermore, $\mathbf{t}$ in (\ref{equ:18}) is decoupled from $\mathbf{R}$, and (\ref{equ:18}) is linear in $\mathbf{t}$. 


Now we consider the least-squares problem (\ref{equ:4}) for the new unknowns. We stack the residuals from  ${{n}_{\pi }}$ point-to-plane, ${{n}_{l}}$ point-to-line, and ${{n}_{p}}$ point-to-point correspondences to get
\begin{equation} \label{equ:19}
{{\bf{e}}_g} = {\bf{Au}} + {\bf{Bt}}
\end{equation}
where $\bf{A}$ is a $\left( {{n}_{\pi }}+3{{n}_{l}}+3{{n}_{p}} \right)\times 11$ matrix and $\bf{B}$ is a $\left( {{n}_{\pi }}+3{{n}_{l}}+3{{n}_{p}} \right)\times 3$ matrix.
Then (\ref{equ:4}) can be written as 
\begin{equation} \label{equ:20}
{C_{\pi lp}}\left(\rho,\alpha,\beta,\gamma,{\bf{t}}\right) = {\bf{e}}_g^T{{\bf{e}}_g}
\end{equation}
As (\ref{equ:19}) is linear for $\mathbf{t}$, then $\bf{t}$ has a closed-form solution as
\begin{equation} \label{equ:21}
\mathbf{t}=-\left( {{\mathbf{B}}^{T}}\mathbf{B} \right)^{-1}{{\mathbf{B}}^{T}}\mathbf{Au}
\end{equation} 
Substituting (\ref{equ:21}) into (\ref{equ:20}), we derive a cost function only involving $\rho, \alpha, \beta, \gamma$ as
\begin{equation} \label{equ:22}
{{C}_{\pi lp}}\left( \rho, \alpha, \beta, \gamma \right)={{\mathbf{u}}^{T}}{{\mathbf{C}}^{T}}\mathbf{Cu}={{\mathbf{u}}^{T}}\mathbf{Qu}
\end{equation}
where ${\bf{C}}=  {{\bf{A}} - {\bf{B}}{{\left( {{{\bf{B}}^T}{\bf{B}}} \right)}^{ - 1}}{{\bf{B}}^T}{\bf{A}}} $.
The elements in the vector $\mathbf{u}$ are second order monomials except for the constant term. Thus (\ref{equ:22}) is a fourth order polynomial function. We compute the first order optimality conditions to get all the stationary points. This first-order condition contains four third order polynomials for  $\rho ,\alpha ,\beta $ and  $\gamma$ as
\begin{equation} \label{equ:23}
\begin{split}
{g_\rho } = \frac{{\partial {C_{\pi lp}}}}{{\partial \rho }} = 0, \quad {g_\alpha } = \frac{{\partial {C_{\pi lp}}}}{{\partial \alpha }} = 0, \\
{g_\beta } = \frac{{\partial {C_{\pi lp}}}}{{\partial \beta }} = 0, \quad {g_\gamma } = \frac{{\partial {C_{\pi lp}}}}{{\partial \gamma }} = 0.
\end{split}
\end{equation} 

According to B\'ezout's theorem \cite{cox2006using}, there exist at most ${{3}^{4}}=81$ solutions. We find that the polynomial system in (\ref{equ:23}) only contains third and first degree monomials. This equation system stratifies the 2-fold symmetry \cite{larsson2016uncovering,ask2012exploiting}. 
That is to say if any nontrivial $\mathbf{\xi}={{\left[ \rho ,\alpha ,\beta ,\gamma  \right]}^{T}}$ is a solution,  $-\mathbf{\xi}$ is also a solution. Thus there are at most 40 independent solutions. We adopt the algorithm introduced in \cite{larsson2017efficient} to generate the polynomial solver which can  utilize this property to yield an efficient solution.

After solving (\ref{equ:23}), we are then able to compute $\mathbf{t}$ from (\ref{equ:21}). The CGR parameters ${{s}_{1}},{{s}_{2}}$ and ${{s}_{3}}$ can be recovered from the definitions in (\ref{equ:16}) and (\ref{equ:17}).
The above formulation treats $\rho ,\alpha ,\beta ,\gamma $ as independent unknowns. However, they are related as they are functions of ${{s}_{1}},{{s}_{2}}$ and ${{s}_{3}}$. 
To recover the minimizer of (\ref{equ:12}), we refine the solution using the Newton-Raphson iteration method in (\ref{equ:15}). We summarize our least-squares solution in Algorithm \ref{alg:LS}.
\begin{algorithm}
	\caption{Least-squares solution} \label{alg:LS}
	\textbf{Input:}  ${{n}_{\pi }}$  point-to-plane, ${{n}_{l}}$ point-to-line, and ${{n}_{p}}$ point-to-point correspondences\\
	\textbf{Output:} $\bf{R}$ and $\bf{t}$ \\
	\begin{tabular}{ll}
		1. & Compute the coefficient matrices $\bf{A}$ and $\bf{B}$ in (\ref{equ:19}). \\
		2. & Compute $\bf{Q}$ in (\ref{equ:22}).\\
		3. & Compute the first order optimal conditions (\ref{equ:23}) \\
		4. & Solve the equation system (\ref{equ:23}) for $\rho ,\alpha ,\beta ,\gamma$.\\
		5. & Recover $s_1, s_2, s_3$ using (\ref{equ:17}). \\
		6. & Compute $\bf{t}$ from (\ref{equ:21}).\\
		7. & Refine the solution by Newton-Raphson iteration (\ref{equ:15}).\\
	\end{tabular}	
\end{algorithm} 	

\section{Minimal Solution}\label{sec:min}

Unlike most pose estimation problems having an unique minimal configuration, our problem has multiple minimal configurations. Specifically, any combination of 3D correspondences with 6 effective constraints forms a minimal configuration. Table \ref{table:mini_config} lists the minimal configurations.  We use the pattern Pt$i$L$j$Pl$k$ to represent $i$  point-to-point, $j$ point-to-line and $k$ point-to-plane correspondences. As one point-to-point, point-to-line and point-to-plane correspondence provide 3, 2 and 1 constraints, respectively, it is easy to verify that the configurations in Table \ref{table:mini_config} are minimal configurations, except for the case Pt$2$L$0$Pl$1$ (\ie 2 point-to-point and 1 point-to-plane correspondences). It seems that Pt$2$L$0$Pl$1$ has 7 constraints, as 2 point-to-point correspondences seemingly provide 6 constraints and 1 point-to-plane correspondence provides 1 constraint. However, the collinear configuration is  degenerate for point-to-point correspondences \cite{arun1987least}. Two point-to-point correspondences form a collinear configuration. One additional constraint is required to recover $\bf{R}$ and $\bf{t}$.  We propose an unified solution for these minimal cases. 
%

\begin{table} 
	\centering
	\begin{tabular}{c|c|c|c}
		\hline
		name &  \#point & \#line &  \#plane \\
		\hline \hline
		Pt$0$L$0$Pl$6$ &	0 &	0 &	6 \\
		Pt$0$L$1$Pl$4$ &	0 &	1 &	4 \\
		Pt$1$L$0$Pl$3$ &	1 &	0 &	3 \\
		Pt$0$L$2$Pl$2$ &	0 &	2 &	2 \\
		Pt$1$L$1$Pl$1$ &	1 &	1 &	1 \\
		Pt$2$L$0$Pl$1$ &	2 &	0 &	1 \\
		Pt$0$L$3$Pl$0$ &	0 &	3 &	0 \\
		\hline		
	\end{tabular}	
	\caption{Minimal configurations.} \label{table:mini_config}		
\end{table}

Let us first consider the rational equation from (\ref{equ:9}). Its numerator, \ie ${\bf{k}}^T{\bf{v}}$, is a third order polynomial.   Assume that we have 6 equations in the form of (\ref{equ:9}). \textit{This will lead to an equation system with 6 third order equations, which is hard to solve.}   We can not use the intermediate unknowns defined in (\ref{equ:16}) and (\ref{equ:17}) to the minimal problem, since this will result in 7 unknowns, but there are only 6 constraints. Here we introduce 3 new intermediate unknowns as
\begin{equation} \label{equ:24}
y_i = \left(1+{\bf{s}}^T{\bf{s}}\right)t_i, i=1,2,3.
\end{equation}
Define ${{\bf{y}}=\left[y_1,y_2,y_3\right]^T}$. Then  (\ref{equ:9}) can be rewritten as
\begin{equation} \label{equ:25}
{r_g} =\frac{{{{\bf{c}}^T}{\bf{\bf{x}}} + {{\bf{a}}^T}{\bf{y}}}}{{1 + {{\bf{s}}^T}{\bf{s}}}} = 0,
\end{equation}
where ${\bf{x}}=\left[s_1^2,s_2^2,s_2^2,s_1s_2,s_1s_3,s_2s_3,s_1,s_2,s_3,1\right]^T$. Define ${\bar r_g}$ the numerator of (\ref{equ:25}). Eliminating the denominator of (\ref{equ:25}), we have
\begin{equation} \label{equ:26}
{\bar r_g ={{{\bf{c}}^T}{\bf{\bf{x}}} + {{\bf{a}}^T}{\bf{y}}}= 0}. 
\end{equation} 
This is a quadratic equation in $s_1, s_2, s_3$ and a linear equation in the new unknowns $y_1, y_2, y_3$. Compared to the original equation from the numerator of (\ref{equ:9}), the new one is much simpler.  For the minimal problem, we have 6 effective equations as (\ref{equ:26}). Let us divide the 6 equations into  two groups. Each group has 3 equations as
\begin{equation} \label{equ:27}
{\bf{C}}_1{\bf{x}}+ {\bf{A}}_1{\bf{y}} = {\bf{0}_{3 \times 1}}, \quad
{\bf{C}}_2{\bf{x}}+ {\bf{A}}_2{\bf{y}} = {\bf{0}_{3 \times 1}},
\end{equation}
We can easily calculate $\bf{y}$ using the first three equations as
\begin{equation} \label{equ:28}
{\bf{y}} = -{\bf{A}}_1^{-1}{\bf{C}}_1{\bf{x}}.
\end{equation}
Substituting (\ref{equ:28}) into the second part of (\ref{equ:27}), we have
\begin{equation} \label{equ:29}
\begin{split}
f_i = &k_{i1}s_1^2+k_{i2}s_2^2+k_{i3}s_3^2+k_{i4}s_1s_2+k_{i5}s_1s_3+\\
&k_{i6}s_2s_3+k_{i7}s_1+k_{i8}s_2+k_{i9}s_3+k_{i10}=0
\end{split}	
\end{equation}
where $k_{ij}$ ($ i \in \left[1,3\right], j \in \left[1,10\right] $) is an element of ${\bf{K}} = {\bf{C}}_2 - {\bf{A}}_2{\bf{A}}_1^{-1}{\bf{C}}_1$. 

\subsection{Hidden Variable Polynomial Solver} \label{subsec:hidden}
Here we present a hidden variable method \cite{cox2006using} to solve (\ref{equ:29}). We treat one unknown as a constant. This unknown is called the hidden variable. Without loss of generality, we treat $s_1$   and $s_2$  as unknowns, and $s_3$  as a constant. Thus (\ref{equ:29}) can be rewritten as an equation system in $s_1$   and $s_2$ as
\begin{equation} \label{equ:30}
\begin{split}
{f_i}  = & {k_{i1}}s_1^2+{k_{i2}}s_2^2+{k_{i4}}{s_1}{s_2}+{p_{i1}}({s_3})s_1+ \\
&{p_{i2}}({s_3}){s_2}+{p_{i3}}({s_3}), i=1,2,3,
\end{split}
\end{equation}
where ${p_{i1}}({s_3})={k_{i5}}{s_3}+{k_{i7}}$, ${p_{i2}}({s_3})={k_{i6}}{s_3}+{k_{i8}}$, ${p_{i3}}({s_3})={k_{i3}}{s_3}^2+{k_{i9}}{s_3}+{k_{i10}}$.  Then we introduce an auxiliary variable ${s_0}$ to convert ${f_i}$ to a homogeneous equation ${F_i}$, \ie, every monomial in $F_i$ is of degree 2. This leads to  the  following homogeneous equation system:
\begin{equation} \label{equ:31}
\begin{split}
{F_i}  = & {k_{i1}}s_1^2+{k_{i2}}s_2^2+{k_{i4}}{s_1}{s_2}+{p_{i1}}({s_3}){s_0}s_1+ \\
&{p_{i2}}({s_3}){s_0}{s_2}+{p_{i3}}({s_3}){s_0^2}=0, i=1,2,3.
\end{split}
\end{equation}
If we treat $s_0^2$, $s_1$ and $s_2$ as variables, we can rewrite (\ref{equ:31}) as a linear homogeneous system in $s_0^2$, $s_1$ and $s_2$ as follows
\begin{equation}  \label{equ:32}
\left[ {\begin{array}{*{20}{c}}
	{{P_{11}}}&{{P_{12}}}&{{P_{13}}}\\
	{{P_{21}}}&{{P_{22}}}&{{P_{23}}}\\
	{{P_{31}}}&{{P_{32}}}&{{P_{33}}}
	\end{array}} \right]\left[ \begin{array}{l}
s_0^2\\
{s_1}\\
{s_2}
\end{array} \right] = {{\bf{0}}_{3 \times 1}},
\end{equation}
where $P_{i1}=p_{i3}(s_3)$, $P_{i2}=k_{i1}{s_1} + k_{i4}{s_2} + p_{i1}(s_3)s_0$, $P_{i3}=k_{i2}s_2 + p_{i2}(s_3)$, $i=1,2,3$. Let us denote the coefficient matrix of (\ref{equ:32}) as ${\bf{P}}=\left(P_{ij}\right)_{3 \times 3}$. If the polynomial system (\ref{equ:31}) has a nontrivial solution, there will exist a nontrivial solution for the  homogeneous linear  system (\ref{equ:32}). According to the algebraic theory, (\ref{equ:32}) has a nontrivial solution if and only if the determinant of  $\bf{P}$ is zero, \ie, $ \det \left( {\bf{P}} \right) = 0$. Define $F_4 = \det \left( {\bf{P}} \right)$. $F_4$ has the same form as $F_i$ in (\ref{equ:31}). Similarly, we can construct two homogeneous linear systems for  $\left[s_0, s_1^2, s_2\right]^T$ and $\left[s_0, s_1, s_2^2\right]^T$. This will yield two homogeneous equations $F_5=0$ and $F_6=0$, respectively. 
Stacking $F_i$ ($i=1,\cdots, 6$) together, we have a homogeneous linear system as
\begin{equation} \label{equ:33}
{\bf{C}}\left(s_3\right){\bf{h} }= {\bf{0}}_{6 \times 1}
\end{equation}
where ${\bf{C}}\left(s_3\right)$ is a $6 \times 6$ matrix with polynomials in $s_3$ as elements, and  ${\bf{h}} = [s_0^2, s_1^2, s_2^2, s_0s_1,  s_0s_2, s_1s_2]^T$. If (\ref{equ:33}) has a nontrivial solution, if and only if $\det \left({\bf{C}}\left(s_3\right)\right) = 0$. This is an eighth order polynomial equation in $s_3$. We can get $s_3$ by solving this equation. Comparing (\ref{equ:30}) and (\ref{equ:31}), we can find that if $[s_1, s_2]$ is a solution of (\ref{equ:30}), $[s_1, s_2, 1]$   is  a solution of (\ref{equ:31}), and vice versa. After we get the solution of $s_3$, we can back-substitute it into  (\ref{equ:33}) and set $s_0 = 1$, then solve the resulting linear system for $s_1$ and $s_2$. After we get $s_1, s_2, s_3$, we can recover $\bf{R}$  by (\ref{equ:5}) and calculate $\bf{y}$ using (\ref{equ:28}). Then $\bf{t}$ can be recovered from (\ref{equ:24}). We summarize our minimal solution in Algorithm \ref{alg:Min}.

\begin{algorithm}
	\caption{Minimal solution} \label{alg:Min}
	\textbf{Input:}  a minimal configuration\\
	\textbf{Output:} $\bf{R}$ and $\bf{t}$ \\
	\begin{tabular}{lp{7.6cm} }
		1. & Compute the coefficient matrices in (\ref{equ:27}). \\
		2. & Compute the coefficients in (\ref{equ:29}). \\
		3. & Solve (\ref{equ:29}) for $s_1,s_2,s_3$ by the hidden variable method described in section \ref{subsec:hidden} and recover $\bf{R}$ using (\ref{equ:5}).\\
		4. & Compute $\bf{y}$ using (\ref{equ:28}) and recover $\bf{t}$ by (\ref{equ:24}).\\
	\end{tabular}		
\end{algorithm} 

\section{Experiments}

In this section, we compare our algorithm with the state-of-the-art algorithms, including Briales's algorithm \cite{briales2017convex}, BnB \cite{olsson2009branch},  Olsson's algorithm \cite{olsson2008solving} and Wientapper's algorithm \cite{wientapper2018universal}. We also consider the  point-to-point case alone, as it has closed-form solution. We adopt Arun's algorithm \cite{arun1987least} implemented in OpenGV \cite{kneip2014opengv}. The algorithms are evaluated in terms of accuracy and computational time using synthetic and real data. 

\begin{table*}
	\small
	\begin{center}
		\begin{tabular}{|l|p{1.3cm}|p{1.3cm}|p{1.2cm}|p{1.3cm}|p{1.3cm}| p{1.2cm}| p{1.3cm}| p{1.3cm}| p{1.2cm}|}
			\hline
			\multirow{2}{*}{Method} &  \multicolumn{3}{|c|}{KITTI 03} & \multicolumn{3}{|c|}{KITTI 04} & \multicolumn{3}{|c|}{KITTI 07}\\
			\cline{2-10}
			& $\bf{t}$ Err ($\%$) & $\bf{R}$ Err ($^\circ$) & Time (s) & $\bf{t}$ Err ($\%$) & $\bf{R}$ Err ($^\circ$) & Time (s) & $\bf{t}$ Err ($\%$) & $\bf{R}$ Err ($^\circ$) & Time (s)  \\
			\hline\hline
			Olsson \cite{olsson2008solving} & 2.03e-1 & \textbf{5.46e-1} & 2.22 & 5.75e-1 & 5.76e-1 & 1.88 & \textbf{3.60e-1} & \textbf{2.53e-1} & 2.08\\
			\hline
			Wientapper \cite{wientapper2018universal} & 2.00e-1 & 5.72e-1 & 0.58 & 4.32e-1 & 3.49e-1 & 0.55 & 3.61e-1 & 2.58e-1 & 0.56\\
			\hline
			Briales \cite{briales2017convex} & 1.96e-1 & 5.61e-1 & 2.16 & \textbf{4.31e-1} & \textbf{3.37e-1} & 1.88 & \textbf{3.60e-1} & \textbf{2.53e-1} & 2.03\\
			\hline
			\textbf{Ours} & \textbf{1.95e-1} & 5.61e-1 & \textbf{0.33} & \textbf{4.31e-1} &  \textbf{3.37e-1} & \textbf{0.30} & \textbf{3.60e-1} & \textbf{2.53e-1} & \textbf{0.30}\\
			\hline
		\end{tabular}
	\end{center}
	\caption{Experimental results on the 03, 04, and 07 sequences of the KITTI dataset \cite{Geiger2013IJRR}. }
	\label{table:accuracy}
\end{table*}

\subsection{Experiments with Synthetic Data}

Our synthetic data is generated as \cite{briales2017convex}. Specifically, each geometric element is determined by randomly sampling a point within a sphere of radius $10m$. For lines and planes, a random unit direction and normal are generated.  We uniformly sample the Euler angles $\alpha ,\beta ,\gamma $ of the rotation matrix ( $\alpha ,\gamma  \in \left[ {0^\circ ,360^\circ } \right]$  and $\beta  \in \left[ {0^\circ ,180^\circ } \right]$). The translation elements are uniformly distributed within $\left[ -10m, 10m\right]$. We use ${\mathbf{\hat R}}$  and ${\mathbf{\hat t}}$ to represent the estimated rotation and translation and use ${{\mathbf{R}}_{gt}}$  and ${{\mathbf{t}}_{gt}}$ to represent the ground truth. The rotation error is evaluated by the angle of the axis-angle representation of ${\mathbf{R}}_{gt}^{ - 1}{\mathbf{R}}$, and the translation error by $\text{ }{{{\left\| {{\mathbf{t}}_{gt}}-\mathbf{\hat{t}} \right\|}_{2}}}/{\left\| {{\mathbf{t}}_{gt}} \right\|}\;$.  We consider the effective number of correspondences as \cite{briales2017convex}. Specifically, the effective number of correspondences for ${{n}_{\pi }}$  point-to-plane, ${{n}_{l}}$ point-to-line, and ${{n}_{p}}$ point-to-point correspondences is calculated as $N = {{n}_{\pi }} + 2{{n}_{l}} + 3{{n}_{p}}$. Given an $N$, we randomly generate a combination of ${n}_{\pi }$, ${{n}_{l}}$ and ${{n}_{p}}$ whose effective number of correspondences is $N$.

\begin{table}
	\small
	\begin{center}
		\begin{tabular}{l|c}
			\hline
			Algorithm & time ($ms$) \\
			\hline\hline
			Our least-squares algorithm & 2.95 \\
			DLSSol & 20.0 \\
			Our minimal algorithm &  0.296 \\
			DMinSol & 0.599\\
			\hline			
		\end{tabular}
	\end{center}
	\caption{Computational time. For the least-squares problem, we consider the time to solve (13), which is independent of the number of correspondences.}
	\label{table:time_vs_orig} 
\end{table}

\begin{figure}
	\centering
	\includegraphics[width=0.23\textwidth]{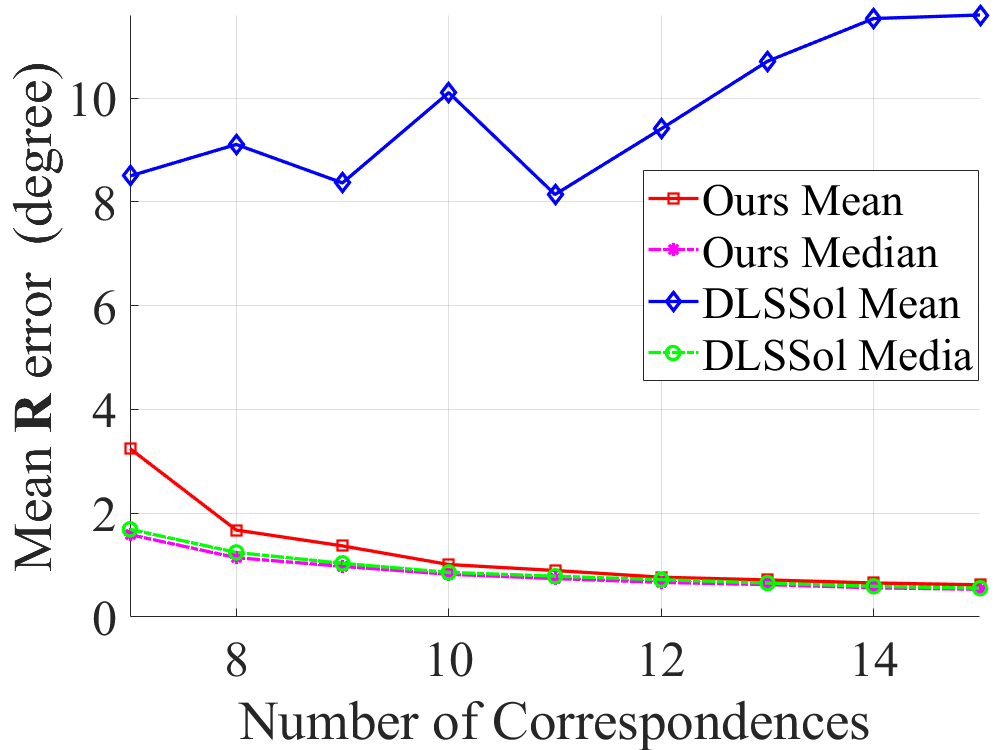}
	\includegraphics[width=0.23\textwidth]{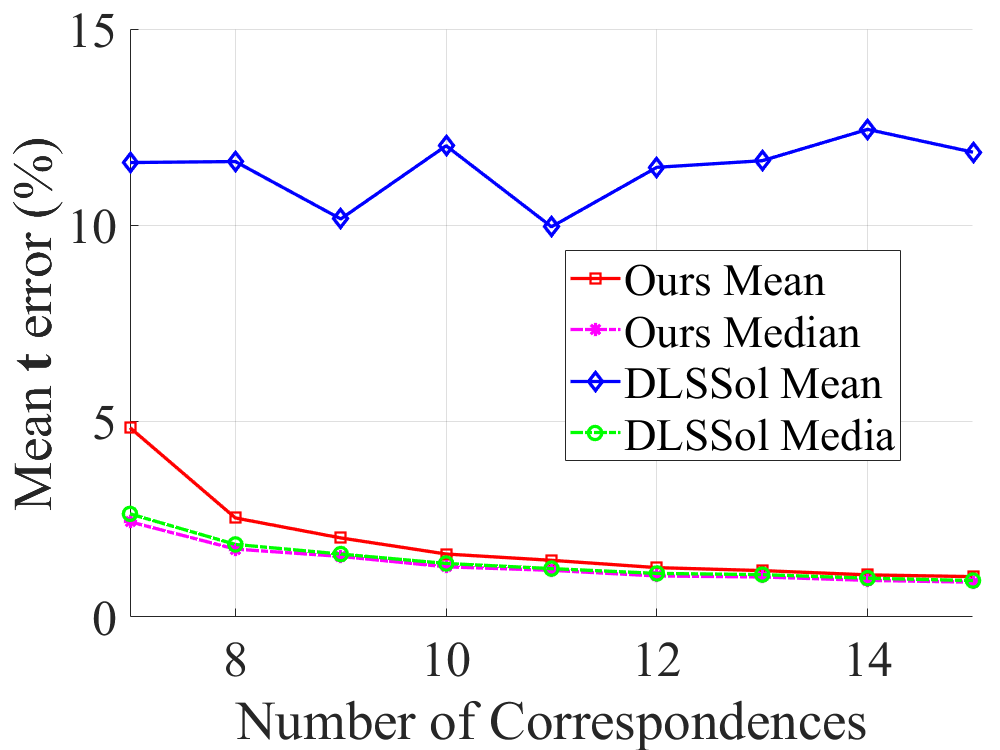}
	\caption{Compare our least-squares solution with direct least-squares solution (DLSSol) which solves the polynomial system (14) using the  Gr\"obner basis method \cite{larsson2017efficient}. }
	\label{fig:vs_DLSSol}
\end{figure}

\begin{figure}
	\centering
	\includegraphics[width=0.23\textwidth]{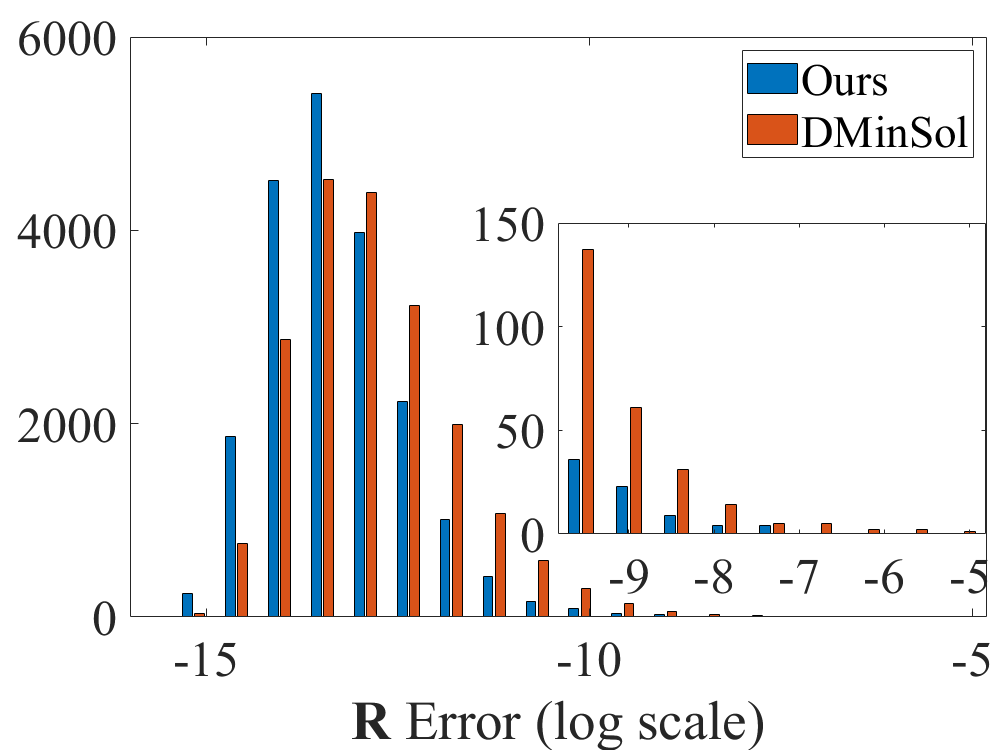}
	\includegraphics[width=0.23\textwidth]{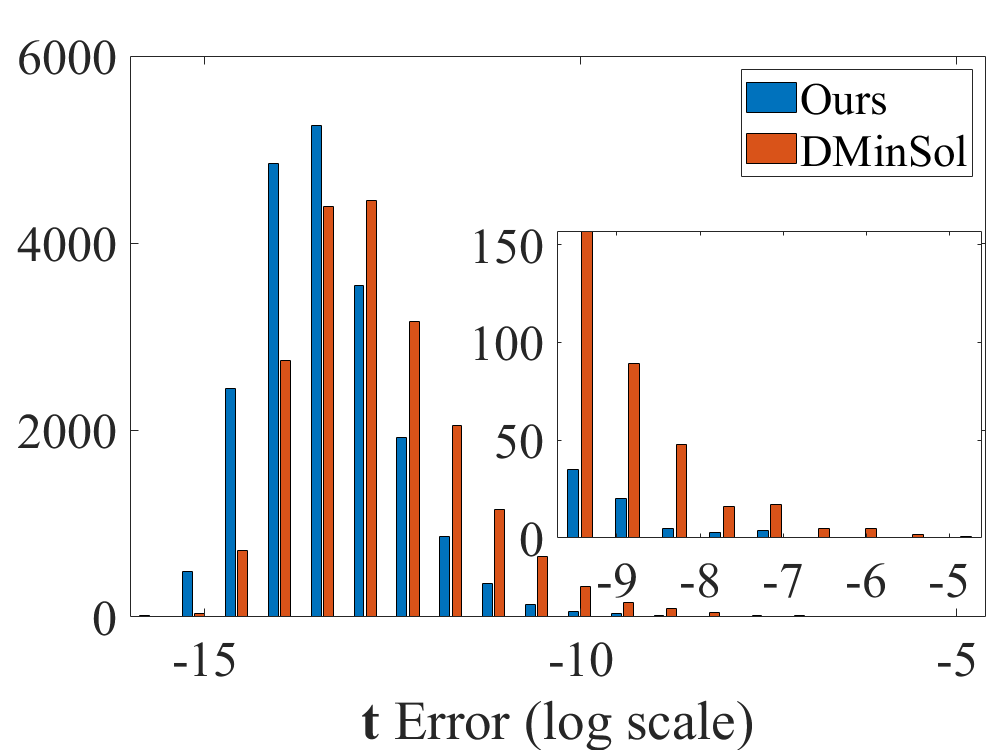}
	\caption{Compare our minimal solution with direct minimal solution (DMinSol) that solves the polynomial system (9) using the Gr\"obner basis method \cite{larsson2017efficient}. }
	\label{fig:vs_DMinSol}
\end{figure}

\noindent
\textbf{Effect of intermediate unknowns} \quad  We introduce intermediate unknowns to simplify the least-squares  and  the minimal problems.  To verify their benefit, we evaluate the performance of the direct least-squares solution (DLSSol) from  solving (13), and the direct minimal solution (DMinSol) from solving the equation system from the numerator of (9) for the minimal problem. We employed the algorithm in \cite{larsson2017efficient} to generate the solvers. For the least-squares problem, we run 2000 trails for each  $N \in [7,15]$. Fig. \ref{fig:vs_DLSSol} shows the results. Directly solving (13) can recover the global minimizer of (4) in theory.  However, the large mean errors of DLSSol verify that this polynomial solver is very unstable. For the minimal problem, we run 20000 trails for  the 6 point-to-plane configuration. Fig. \ref{fig:vs_DMinSol} shows the results. The DMinSol has a much longer tail than our algorithm. 

Table~\ref{table:time_vs_orig} lists the average computational time of different algorithms. For the least-squares problem, we compare the computational time of solving the first order optimality conditions (13). This time is independent of the number of correspondences.  As shown in Table~\ref{table:time_vs_orig}, our least-squares solution is about 7 times faster than DLSSol, and our minimal solution is about 2 times faster than DMinSol. 

The above results verify that our intermediate unknowns can increase the numeric stability as well as reduce the computational time. 

\textbf{Least-squares solution} \quad We conduct experiments to evaluate the performances of different algorithms under varying number of correspondences, increasing level of noise and computational time.   The results of all experiments are from 100 independent trials as \cite{briales2017convex}.  

\begin{figure*}
	\centering
	\includegraphics[width=0.98\textwidth]{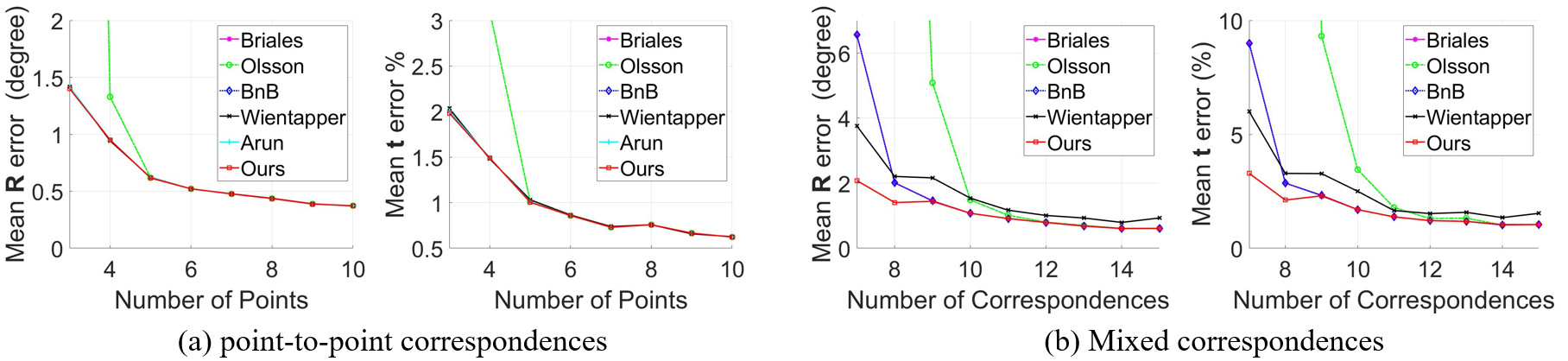}
	\caption{Rotation and translation errors for increasing number of correspondences. The noise level is fixed to 0.05m. }
	\label{fig:vary_num}
\end{figure*}

The first experiment considers a fixed noise level and an increasing number of correspondences. Let us denote the standard deviation of a zero mean Gaussian noise distribution as $\delta $. We set $\delta =0.05m$. $N$ varies from $7$ to $15$. 
Fig. \ref{fig:vary_num} shows the result. For the pose-to-point case, most of the algorithms can recover the optimal solution except for \cite{olsson2008solving} when the number of points is small. For the mixed correspondences, the results of BnB \cite{olsson2009branch}  and Briales's algorithm \cite{briales2017convex} overlap. This is consistent to the results in \cite{briales2017convex}.  Our algorithm significantly outperforms previous works when $N$ is small as the ambiguous case more likely happens  for a small $N$. When $N$ is large, our algorithm achieves the same accuracy as \cite{olsson2009branch} that has guaranteed global optimality.   Wientapper's algorithm \cite{wientapper2018universal} can not find the optimal solution even $N$ is large. 

\begin{figure*}
	\centering
	\includegraphics[width=0.98\textwidth]{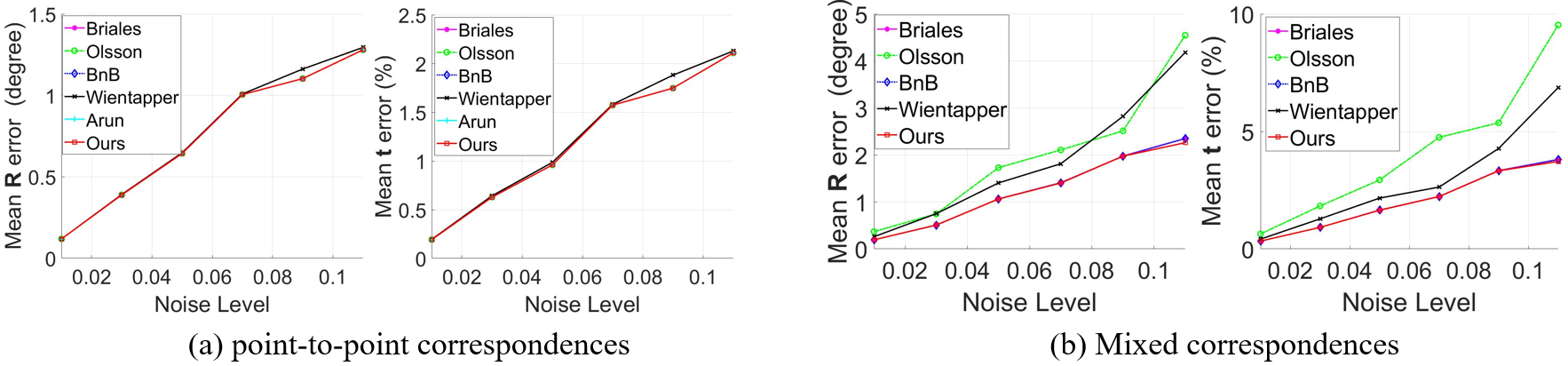}
	\caption{Rotation and translation errors for increasing noise level. The number of point correspondences is fixed to 5 and the number of effective correspondences is set to 10.}
	\label{fig:vary_noise} 
\end{figure*}

In the second experiment,  $\delta$ is from $0.01m$ to $0.11m$, stepping by 0.02m. The results are illustrated in Fig. \ref{fig:vary_noise}.  For the point-to-point case, most of the algorithms provide the optimal solution as \cite{arun1987least}, except for \cite{wientapper2018universal} when $\delta = 0.1m$. For the mixed case,  BnB,  Briales's algorithm and our algorithm present better results than other algorithms. Our algorithm gives a slightly better result than BnB and  Briales's algorithm when $\delta =0.11m$.

\begin{figure}
	\centering
	\includegraphics[width=0.26\textwidth]{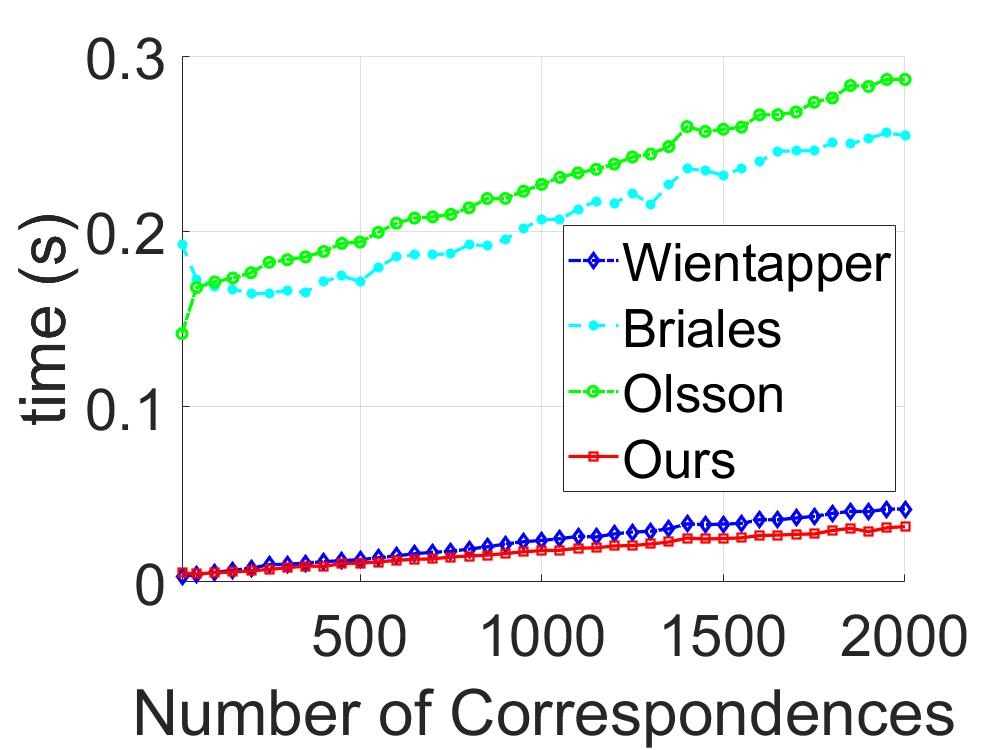}
	\caption{Computational time of different algorithms.}
	\label{fig:time}
\end{figure}

In the last experiment, we compare the computational time of different algorithms. We vary the effective number of correspondences $N$ from 10 to 2000. For every $N$, we run each algorithm 100 times to calculate the average running time.  Fig. \ref{fig:time} provides the result. We did not run the BnB algorithm \cite{olsson2009branch}, because it is too slow for a large $N$. It is clear that our algorithm is the fastest one among the compared algorithms. Wientapper's algorithm \cite{wientapper2018universal} is efficient. 
But  their algorithm needs to compute a problem 4 times, thus the gap between the computational time of two algorithms increases as $N$ enlarges. Actually, $N$ could be 10 times larger in real applications than our simulation, as shown in Table \ref{table:n-corres}. 

\begin{figure}
	\centering
	\includegraphics[width=0.46\textwidth]{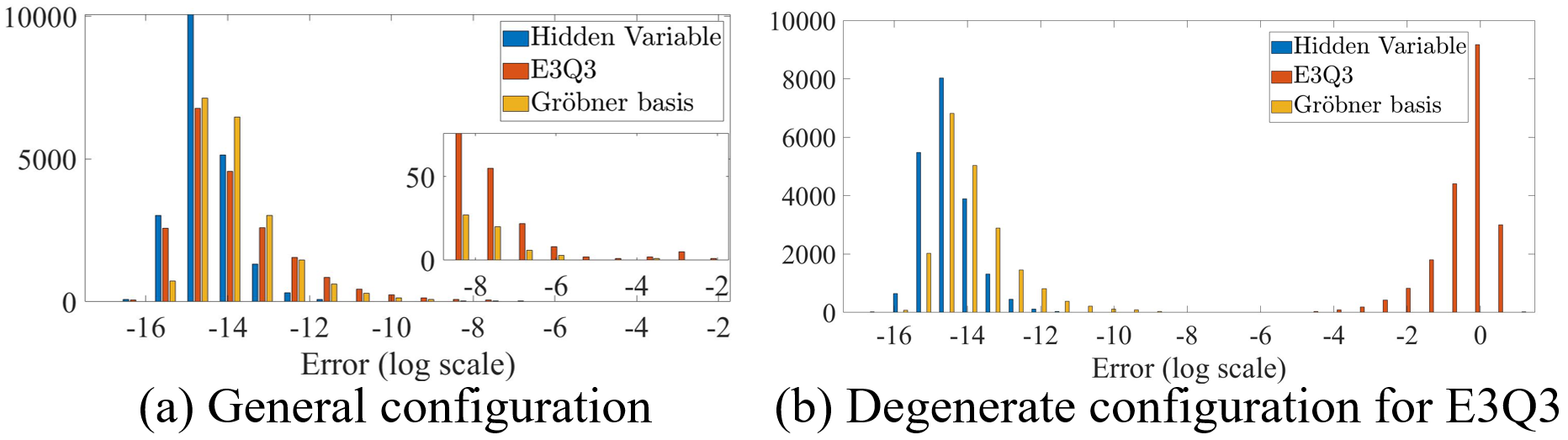}
	\caption{Numeric stability of polynomial solvers for (\ref{equ:29}).}
	\label{fig:poly}
\end{figure}

\textbf{Minimal solution} \quad The equation system (\ref{equ:29}) is important for the estimation of $\bf{R}$. Actually, Ramalingam \etal \cite{ramalingam2013theory} formulated some point-to-plane configurations as a quadratic equation system as (\ref{equ:29}).  We first consider the numeric stability of the polynomial solver for  (\ref{equ:29}). We compare our hidden variable method, E3Q3 \cite{kukelova2016efficient}, and the Gr\"obner basis polynomial solver generated by \cite{larsson2017efficient}. We randomly  generate a real solution and the coefficients of (\ref{equ:29}). Then we substitute this real solution into  (\ref{equ:29}) to calculate the constant terms of (\ref{equ:29}).  We run each algorithm 20000 trails and compute  the estimation error of this real solution. Fig. \ref{fig:poly} (a) shows  the estimation error histograms of the compared algorithms. It is clear that our hidden variable method is more stable than other algorithms. The histograms of E3Q3 and the Gr\"obner basis method have long tails. This is because E3Q3 and the Gr\"obner basis method require to compute the inverse of a matrix. When a matrix approximates a singular matrix, the performance of these algorithms will degrade.  Fig. \ref{fig:poly} (b) demonstrates the performance of E3Q3 under a degenerate case. We generate nearly singular matrix for E3Q3, whose least singular value is in $(0,10^{-6})$ . In this situation, E3Q3 performs very bad. 

\begin{figure}
	\centering
	\includegraphics[width=0.23\textwidth]{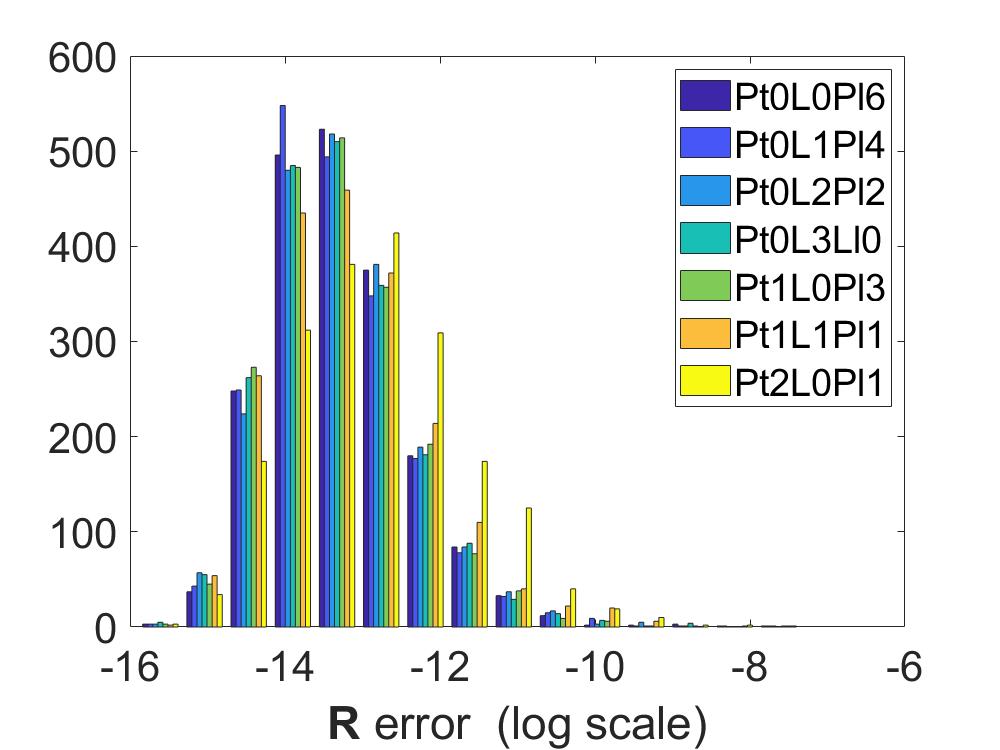}
	\includegraphics[width=0.23\textwidth]{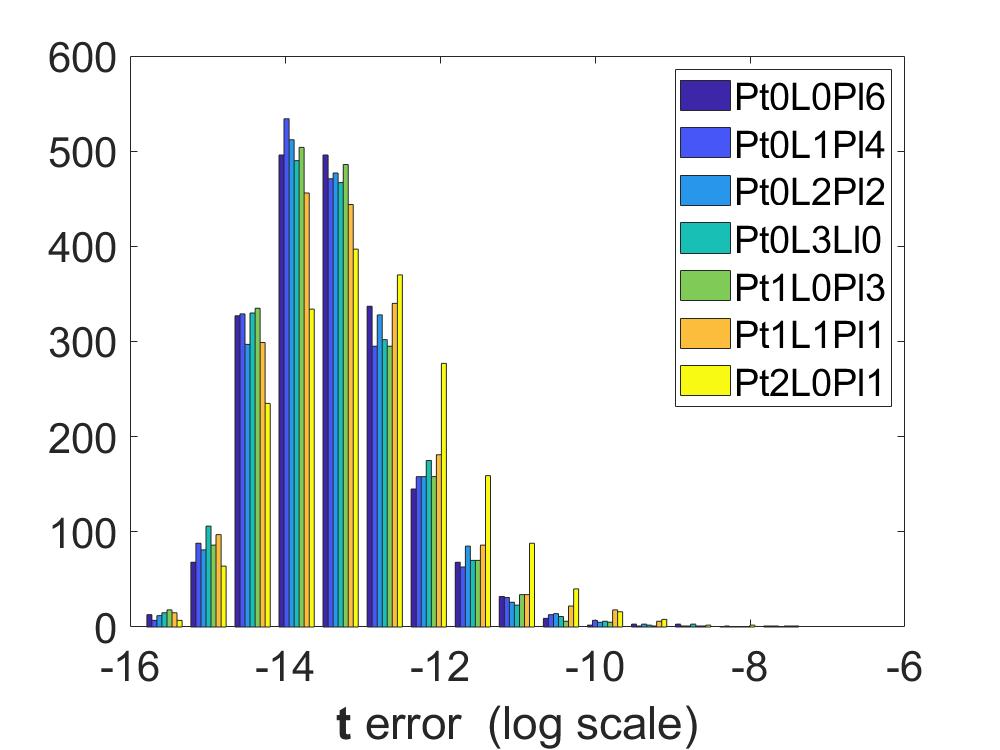}
	\caption{Numeric stability of our minimal solution.}
	\label{fig:mini}
\end{figure}

We also evaluate the numeric stability of our unified minimal solution. We run 2000 independent trails for each minimal configuration.  As demonstrated in Fig. \ref{fig:mini}, our algorithm provides accurate solutions for all minimal configurations. This avoids solving each configuration case by case.

\begin{table}
	\small
	\begin{center}
		\begin{tabular}{|l|c|c|c|}
			\hline
			\multirow{2}{*}{Sequences} & \multicolumn{3}{|c|}{Average Correspondences Per Frame}\\
			\cline{2-4}
			& \#P-to-P & \#P-to-L & \#P-to-PL\\
			\hline\hline
			KITTI 03 & 63 & 17 & 21117\\
			\hline
			KITTI 04 & 37 & 30 & 19994 \\ 
			\hline
			KITTI 07 & 111 & 40 & 20126\\
			\hline
		\end{tabular}
	\end{center}
	\caption{Average numbers of point-to-point(P-to-P), point-to-pine(P-to-L), and point-to-plane(P-to-PL) correspondences per frame of 03, 04 and 07 KITTI sequences.} 	\label{table:n-corres} 
\end{table}

\subsection{Experiments with Real Data}

We generated our real-world dataset from the KITTI  dataset \cite{Geiger2013IJRR}. Our dataset contains point-to-point, point-to-line and point-to-plane correspondences described in Table \ref{table:n-corres}. There are more than 20000 correspondences in each frame, with the majority of them being point-to-plane correspondences. We evaluate the accuracy and running time of different algorithms on this dataset. 



For each frame, 2D feature points are detected by the ORB feature detector \cite{Rublee2011ICCV}. We project LiDAR points into the image plane, and select  the LiDAR points  around an ORB feature. Then, we fit a local plane to these LiDAR points. Finally, we calculate the 3D coordinates of an ORB feature by calculating the intersection of the back-projection ray of the ORB feature and the local plane. We use the ORB descriptors to match 2D feature points, and obtain the 3D point-to-point correspondences.


Next, we generate  point-to-line correspondences. For each frame, 2D lines are detected by  the Line Segment Detector (LSD) \cite{Gioi2010TPAMI} and described by the Line Band Descriptor (LBD) \cite{zhang2013efficient}.  A 2D line is represented by two 2D endpoints. The 3D endpoints of a line is generated as the 2D feature points described above. We then generate 3D line correspondences by matching  their LBD features. Given a line-to-line correspondence, two point-to-line correspondences are generated for each of its endpoints.


Finally, we calculated  point-to-plane correspondences. For each frame, planes are extracted from  LiDAR points by the region growing algorithm \cite{poppinga2008fast}. To match a plane point to a previous plane, we find the nearest LiDAR point in previous  planes. 


We evaluate the performance of different algorithms on sequences 03, 04, and 07 of the KITTI dataset. We did not test  BnB \cite{olsson2009branch}, as it is extremely slow on this large dataset. Our algorithm achieves the same or slightly better result as the  state-of-the-art algorithm \cite{briales2017convex} while being around 7 times faster, as shown in Table ~\ref{table:accuracy}.

\section{Conclusions}
In this paper, we present an efficient and accurate least-squares solution and an unified minimal solution for the 3D registration problem.  We proof that there exist ambiguous configurations for any number of point-to-plane and point-to-line correspondences. This requires an algorithm has the ability to reveal local minimizers.  We use the CGR parameterization to represent the rotation, removing the quadratic constraints on the rotation. However, this results in a rational form residual which is hard to solve. We introduce several intermediate variables to simplify the first-order optimality conditions of the  least-squares problem, and the equation system of the minimal configuration. We evaluate our algorithm through synthetic and real data. The experimental results show that computing local minimizers is essential, especially when  $N$ is small. Besides, our algorithm is as accurate as previous globally optimal solutions when $N$ is large, but is much faster.

{\small
 \bibliographystyle{ieee}
 \bibliography{plp_registration_arxiv}

\begin{thebibliography}{10}\itemsep=-1pt

\bibitem{arun1987least}
K.~S. Arun, T.~S. Huang, and S.~D. Blostein.
\newblock Least-squares fitting of two 3-d point sets.
\newblock {\em IEEE Transactions on pattern analysis and machine intelligence},
  (5):698--700, 1987.

\bibitem{ask2012exploiting}
E.~Ask, Y.~Kuang, and K.~{\AA}str{\"o}m.
\newblock Exploiting p-fold symmetries for faster polynomial equation solving.
\newblock In {\em Proceedings of the 21st International Conference on Pattern
  Recognition (ICPR2012)}, pages 3232--3235. IEEE, 2012.

\bibitem{besl1992method}
P.~J. Besl and N.~D. McKay.
\newblock Method for registration of 3-d shapes.
\newblock In {\em Sensor Fusion IV: Control Paradigms and Data Structures},
  volume 1611, pages 586--607. International Society for Optics and Photonics,
  1992.

\bibitem{briales2017convex}
J.~Briales, J.~Gonzalez-Jimenez, et~al.
\newblock Convex global 3d registration with lagrangian duality.
\newblock In {\em International Conference on Computer Vision and Pattern
  Recognition (CVPR)}, 2017.

\bibitem{censi2008icp}
A.~Censi.
\newblock An icp variant using a point-to-line metric.
\newblock In {\em Robotics and Automation, 2008. ICRA 2008. IEEE International
  Conference on}, pages 19--25. IEEE, 2008.

\bibitem{chen1990pose}
H.~H. Chen.
\newblock Pose determination from line-to-plane correspondences: Existence
  condition and closed-form solutions.
\newblock In {\em [1990] Proceedings Third International Conference on Computer
  Vision}, pages 374--378. IEEE, 1990.

\bibitem{87340}
H.~H. Chen.
\newblock Pose determination from line-to-plane correspondences: existence
  condition and closed-form solutions.
\newblock {\em IEEE Transactions on Pattern Analysis and Machine Intelligence},
  13(6):530--541, Jun 1991.

\bibitem{cox2006using}
D.~A. Cox, J.~Little, and D.~O'shea.
\newblock {\em Using algebraic geometry}, volume 185.
\newblock Springer Science \& Business Media, 2006.

\bibitem{fischler1981random}
M.~A. Fischler and R.~C. Bolles.
\newblock Random sample consensus: a paradigm for model fitting with
  applications to image analysis and automated cartography.
\newblock {\em Communications of the ACM}, 24(6):381--395, 1981.

\bibitem{Geiger2013IJRR}
A.~Geiger, P.~Lenz, C.~Stiller, and R.~Urtasun.
\newblock Vision meets robotics: The kitti dataset.
\newblock {\em International Journal of Robotics Research (IJRR)}, 2013.

\bibitem{gomez2015extrinsic}
R.~Gomez-Ojeda, J.~Briales, E.~Fernandez-Moral, and J.~Gonzalez-Jimenez.
\newblock Extrinsic calibration of a 2d laser-rangefinder and a camera based on
  scene corners.
\newblock In {\em 2015 IEEE International Conference on Robotics and Automation
  (ICRA)}, pages 3611--3616. IEEE, 2015.

\bibitem{hesch2011direct}
J.~A. Hesch and S.~I. Roumeliotis.
\newblock A direct least-squares (dls) method for pnp.
\newblock In {\em 2011 International Conference on Computer Vision}, pages
  383--390. IEEE, 2011.

\bibitem{horn1987closed}
B.~K. Horn.
\newblock Closed-form solution of absolute orientation using unit quaternions.
\newblock {\em Josa a}, 4(4):629--642, 1987.

\bibitem{horn1988closed}
B.~K. Horn, H.~M. Hilden, and S.~Negahdaripour.
\newblock Closed-form solution of absolute orientation using orthonormal
  matrices.
\newblock {\em JOSA A}, 5(7):1127--1135, 1988.

\bibitem{kneip2014opengv}
L.~Kneip and P.~Furgale.
\newblock Opengv: A unified and generalized approach to real-time calibrated
  geometric vision.
\newblock In {\em 2014 IEEE International Conference on Robotics and Automation
  (ICRA)}, pages 1--8. IEEE, 2014.

\bibitem{kneip2014upnp}
L.~Kneip, H.~Li, and Y.~Seo.
\newblock Upnp: An optimal o (n) solution to the absolute pose problem with
  universal applicability.
\newblock In {\em European Conference on Computer Vision}, pages 127--142.
  Springer, 2014.

\bibitem{kukelova2016efficient}
Z.~Kukelova, J.~Heller, and A.~Fitzgibbon.
\newblock Efficient intersection of three quadrics and applications in computer
  vision.
\newblock In {\em Proceedings of the IEEE Conference on Computer Vision and
  Pattern Recognition}, pages 1799--1808, 2016.

\bibitem{larsson2016uncovering}
V.~Larsson and K.~{\AA}str{\"o}m.
\newblock Uncovering symmetries in polynomial systems.
\newblock In {\em European Conference on Computer Vision}, pages 252--267.
  Springer, 2016.

\bibitem{larsson2017efficient}
V.~Larsson, K.~Astrom, and M.~Oskarsson.
\newblock Efficient solvers for minimal problems by syzygy-based reduction.
\newblock In {\em Proceedings of the IEEE Conference on Computer Vision and
  Pattern Recognition}, pages 820--829, 2017.

\bibitem{li20073d}
H.~Li and R.~Hartley.
\newblock The 3d-3d registration problem revisited.
\newblock In {\em 2007 IEEE 11th International Conference on Computer Vision},
  pages 1--8. IEEE, 2007.

\bibitem{mirzaei2011optimal}
F.~M. Mirzaei and S.~I. Roumeliotis.
\newblock Optimal estimation of vanishing points in a manhattan world.
\newblock In {\em 2011 International Conference on Computer Vision}, pages
  2454--2461. IEEE, 2011.

\bibitem{naroditsky2011automatic}
O.~Naroditsky, A.~Patterson, and K.~Daniilidis.
\newblock Automatic alignment of a camera with a line scan lidar system.
\newblock In {\em 2011 IEEE International Conference on Robotics and
  Automation}, pages 3429--3434. IEEE, 2011.

\bibitem{newcombe2011kinectfusion}
R.~A. Newcombe, S.~Izadi, O.~Hilliges, D.~Molyneaux, D.~Kim, A.~J. Davison,
  P.~Kohi, J.~Shotton, S.~Hodges, and A.~Fitzgibbon.
\newblock Kinectfusion: Real-time dense surface mapping and tracking.
\newblock In {\em 2011 IEEE International Symposium on Mixed and Augmented
  Reality}, pages 127--136. IEEE, 2011.

\bibitem{olsson2008solving}
C.~Olsson and A.~Eriksson.
\newblock Solving quadratically constrained geometrical problems using
  lagrangian duality.
\newblock In {\em 2008 19th International Conference on Pattern Recognition},
  pages 1--5. IEEE, 2008.

\bibitem{olsson2006registration}
C.~Olsson, F.~Kahl, and M.~Oskarsson.
\newblock The registration problem revisited: Optimal solutions from points,
  lines and planes.
\newblock In {\em Computer Vision and Pattern Recognition, 2006 IEEE Computer
  Society Conference on}, volume~1, pages 1206--1213. IEEE, 2006.

\bibitem{olsson2009branch}
C.~Olsson, F.~Kahl, and M.~Oskarsson.
\newblock Branch-and-bound methods for euclidean registration problems.
\newblock {\em IEEE Transactions on Pattern Analysis and Machine Intelligence},
  31(5):783--794, 2009.

\bibitem{papazov2011stochastic}
C.~Papazov and D.~Burschka.
\newblock Stochastic global optimization for robust point set registration.
\newblock {\em Computer Vision and Image Understanding}, 115(12):1598--1609,
  2011.

\bibitem{poppinga2008fast}
J.~Poppinga, N.~Vaskevicius, A.~Birk, and K.~Pathak.
\newblock Fast plane detection and polygonalization in noisy 3d range images.
\newblock In {\em 2008 IEEE/RSJ International Conference on Intelligent Robots
  and Systems}, pages 3378--3383. IEEE, 2008.

\bibitem{pvribyl2017absolute}
B.~P{\v{r}}ibyl, P.~Zem{\v{c}}{\'\i}k, and M.~{\v{C}}ad{\'\i}k.
\newblock Absolute pose estimation from line correspondences using direct
  linear transformation.
\newblock {\em Computer Vision and Image Understanding}, 161:130--144, 2017.

\bibitem{proencca2018probabilistic}
P.~F. Proen{\c{c}}a and Y.~Gao.
\newblock Probabilistic rgb-d odometry based on points, lines and planes under
  depth uncertainty.
\newblock {\em Robotics and Autonomous Systems}, 104:25--39, 2018.

\bibitem{ramalingam2013theory}
S.~Ramalingam and Y.~Taguchi.
\newblock A theory of minimal 3d point to 3d plane registration and its
  generalization.
\newblock {\em International journal of computer vision}, 102(1-3):73--90,
  2013.

\bibitem{ramalingam2010p2pi}
S.~Ramalingam, Y.~Taguchi, T.~K. Marks, and O.~Tuzel.
\newblock P2$\pi$: A minimal solution for registration of 3d points to 3d
  planes.
\newblock In {\em European Conference on Computer Vision}, pages 436--449.
  Springer, 2010.

\bibitem{Gioi2010TPAMI}
G.~Randall, J.~Jakubowicz, R.~G. von Gioi, and J.~Morel.
\newblock Lsd: A fast line segment detector with a false detection control.
\newblock {\em IEEE Transactions on Pattern Analysis and Machine Intelligence},
  32:722--732, 12 2008.

\bibitem{Rublee2011ICCV}
E.~Rublee, V.~Rabaud, K.~Konolige, and G.~R. Bradski.
\newblock Orb: An efficient alternative to sift or surf.
\newblock 2011.

\bibitem{segal2009generalized}
A.~Segal, D.~Haehnel, and S.~Thrun.
\newblock Generalized-icp.
\newblock In {\em Robotics: science and systems}, volume~2, pages 742--749.
  IEEE, 2015.

\bibitem{serafin2015nicp}
J.~Serafin and G.~Grisetti.
\newblock Nicp: Dense normal based point cloud registration.
\newblock In {\em 2015 IEEE/RSJ International Conference on Intelligent Robots
  and Systems (IROS)}, pages 742--749. IEEE, 2015.

\bibitem{taguchi2013point}
Y.~Taguchi, Y.-D. Jian, S.~Ramalingam, and C.~Feng.
\newblock Point-plane slam for hand-held 3d sensors.
\newblock In {\em 2013 IEEE International Conference on Robotics and
  Automation}, pages 5182--5189. IEEE, 2013.

\bibitem{unnikrishnan2005fast}
R.~Unnikrishnan and M.~Hebert.
\newblock Fast extrinsic calibration of a laser rangefinder to a camera.
\newblock {\em Robotics Institute, Pittsburgh, PA, Tech. Rep. CMU-RI-TR-05-09},
  2005.

\bibitem{vasconcelos2012minimal}
F.~Vasconcelos, J.~P. Barreto, and U.~Nunes.
\newblock A minimal solution for the extrinsic calibration of a camera and a
  laser-rangefinder.
\newblock {\em IEEE transactions on pattern analysis and machine intelligence},
  34(11):2097--2107, 2012.

\bibitem{wientapper2016unifying}
F.~Wientapper and A.~Kuijper.
\newblock Unifying algebraic solvers for scaled euclidean registration from
  point, line and plane constraints.
\newblock In {\em Asian Conference on Computer Vision}, pages 52--66. Springer,
  2016.

\bibitem{wientapper2018universal}
F.~Wientapper, M.~Schmitt, M.~Fraissinet-Tachet, and A.~Kuijper.
\newblock A universal, closed-form approach for absolute pose problems.
\newblock {\em Computer Vision and Image Understanding}, 173:57--75, 2018.

\bibitem{xu2017pose}
C.~Xu, L.~Zhang, L.~Cheng, and R.~Koch.
\newblock Pose estimation from line correspondences: A complete analysis and a
  series of solutions.
\newblock {\em IEEE transactions on pattern analysis and machine intelligence},
  39(6):1209--1222, 2017.

\bibitem{yang2013go}
J.~Yang, H.~Li, and Y.~Jia.
\newblock Go-icp: Solving 3d registration efficiently and globally optimally.
\newblock In {\em Proceedings of the IEEE International Conference on Computer
  Vision}, pages 1457--1464, 2013.

\bibitem{zhang2014loam}
J.~Zhang and S.~Singh.
\newblock Loam: Lidar odometry and mapping in real-time.
\newblock In {\em Robotics: Science and Systems}, volume~2, page~9, 2014.

\bibitem{zhang2015visual}
J.~Zhang and S.~Singh.
\newblock Visual-lidar odometry and mapping: Low-drift, robust, and fast.
\newblock In {\em 2015 IEEE International Conference on Robotics and Automation
  (ICRA)}, pages 2174--2181. IEEE, 2015.

\bibitem{zhang2013efficient}
L.~Zhang and R.~Koch.
\newblock An efficient and robust line segment matching approach based on lbd
  descriptor and pairwise geometric consistency.
\newblock {\em Journal of Visual Communication and Image Representation},
  24(7):794--805, 2013.

\bibitem{zhang2004extrinsic}
Q.~Zhang and R.~Pless.
\newblock Extrinsic calibration of a camera and laser range finder (improves
  camera calibration).
\newblock In {\em 2004 IEEE/RSJ International Conference on Intelligent Robots
  and Systems (IROS)(IEEE Cat. No. 04CH37566)}, volume~3, pages 2301--2306.
  IEEE, 2004.

\bibitem{zhou2014new}
L.~Zhou.
\newblock A new minimal solution for the extrinsic calibration of a 2d lidar
  and a camera using three plane-line correspondences.
\newblock {\em IEEE Sensors Journal}, 14(2):442--454, 2014.

\bibitem{zhou2018automatic}
L.~Zhou, Z.~Li, and M.~Kaess.
\newblock Automatic extrinsic calibration of a camera and a 3d lidar using line
  and plane correspondences.
\newblock In {\em 2018 IEEE/RSJ International Conference on Intelligent Robots
  and Systems (IROS)}, pages 5562--5569. IEEE, 2018.

\bibitem{zhou2016fast}
Q.-Y. Zhou, J.~Park, and V.~Koltun.
\newblock Fast global registration.
\newblock In {\em European Conference on Computer Vision}, pages 766--782.
  Springer, 2016.

\end{thebibliography}
}

\appendix
\section{Proof for  lemma \ref{lemma:line}, lemma \ref{lemma:plane} and theorem \ref{theorm:1} }
As we mentioned in the paper, point-to-line and point-to-plane correspondences have ambiguous configurations. Here we prove lemma \ref{lemma:line}, lemma \ref{lemma:plane} and theorem \ref{theorm:1} in our paper.

	\noindent
	\textbf{Lemma 1.} For any $n_l$ points and any $\left\{ {{{\bf{R}}_1},{{\bf{t}}_1}} \right\}$ and $\left\{ {{{\bf{R}}_2},{{\bf{t}}_2}} \right\}$, there exist  $n_l$ lines to make $\left\{ {{{\bf{R}}_1},{{\bf{t}}_1}} \right\}$ and $\left\{ {{{\bf{R}}_2},{{\bf{t}}_2}} \right\}$ are exact solutions for the $n_l$ point-to-line correspondences.

\begin{figure}	
	\centering	
	\includegraphics[width=0.49\textwidth]{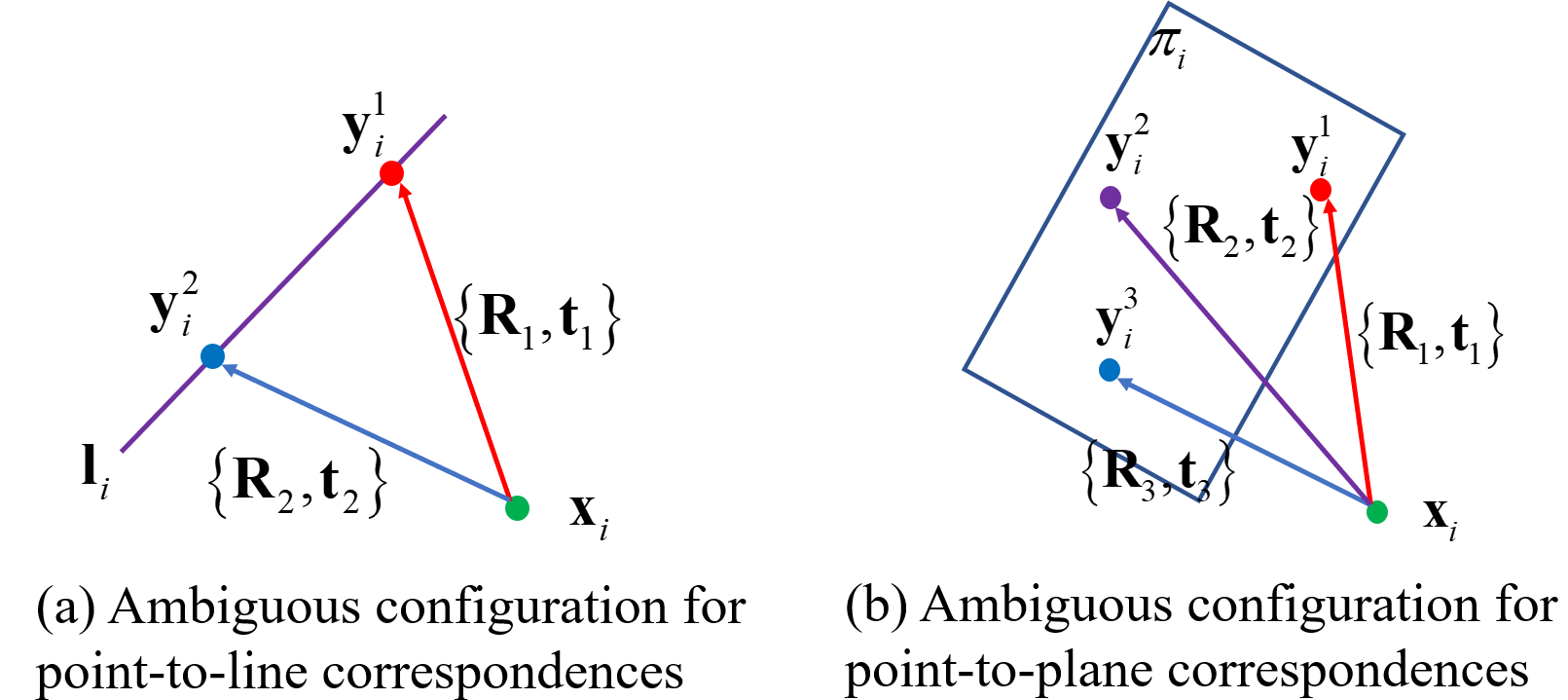}
	\caption{A schematic of point-to-line ambiguous configuration (a) and point-to-plane ambiguous configuration (b). In (a),  $\left\{ {{{\bf{R}}_1},{{\bf{t}}_1}} \right\}$ and $\left\{ {{{\bf{R}}_2},{{\bf{t}}_2}} \right\}$ will transform ${\bf{x}}_i$ to ${\bf{l}}_i$, In (b),  $\left\{ {{{\bf{R}}_1},{{\bf{t}}_1}} \right\}$, $\left\{ {{{\bf{R}}_2},{{\bf{t}}_2}} \right\}$ and $\left\{ {{{\bf{R}}_3},{{\bf{t}}_3}} \right\}$  will transform ${\bf{x}}_i$ to ${\bf{\pi}}_i$} 	\label{fig:ambiguous_l_pl}	
\end{figure}

\begin{proof}
	Define ${{\bf{x}}_i}$   the $i$th   point  of an arbitrary point set with ${n_l}$ points. For any $\left\{ {{{\bf{R}}_1},{{\bf{t}}_1}} \right\}$  and  $\left\{ {{{\bf{R}}_2},{{\bf{t}}_2}} \right\}$, we can have ${\bf{y}}_i^1 = {{\bf{R}}_1}{{\bf{x}}_i} + {{\bf{t}}_1}$
	and  ${\bf{y}}_i^2 = {{\bf{R}}_2}{{\bf{x}}_i} + {{\bf{t}}_2}$. Then we can use ${\bf{y}}_i^1$   and ${\bf{y}}_i^2$   to define a line ${{\bf{l}}_i} = \left[ {{{\bf{d}}_i};{{\bf{y}}_i}} \right]$   with direct ${{\bf{d}}_i} = \frac{{{\bf{y}}_i^1 - {\bf{y}}_i^2}}{{{{\left\| {{\bf{y}}_i^1 - {\bf{y}}_i^2} \right\|}_2}}}$   and a point ${{\bf{y}}_i} = {\bf{y}}_i^1$ on it, as demonstrated in Fig.~\ref{fig:ambiguous_l_pl} (a). According to how we construct ${\bf{l}}_i$, we know that ${\bf{l}}_i$ passes through ${\bf{y}}_i^1$   and ${\bf{y}}_i^2$. Therefore $\left\{ {{{\bf{R}}_1},{{\bf{t}}_1}} \right\}$   and $\left\{ {{{\bf{R}}_2},{{\bf{t}}_2}} \right\}$  are two solutions for the ${n_l}$   point-to-line  $\left\{ {{{\bf{x}}_i} \leftrightarrow {{\bf{l}}_i}} \right\}_{i = 1}^{{n_l}}$.   
\end{proof}

	\noindent
	\textbf{Lemma 2.} For any $n_{\pi}$ points and any $\left\{ {{{\bf{R}}_1},{{\bf{t}}_1}} \right\}$, $\left\{ {{{\bf{R}}_2},{{\bf{t}}_2}} \right\}$ and $\left\{ {{{\bf{R}}_3},{{\bf{t}}_3}} \right\}$, there exist  $n_{\pi}$ planes to make $\left\{ {{{\bf{R}}_1},{{\bf{t}}_1}} \right\}$, $\left\{ {{{\bf{R}}_2},{{\bf{t}}_2}} \right\}$ and $\left\{ {{{\bf{R}}_3},{{\bf{t}}_3}} \right\}$ are exact solutions for the $n_{\pi}$ point-to-plane correspondences.
\begin{proof}
	Define ${{\bf{x}}_i}$   the $i$th   point  of an arbitrary point set with $n_{\pi}$ points. For any $\left\{ {{{\bf{R}}_1},{{\bf{t}}_1}} \right\}$, $\left\{ {{{\bf{R}}_2},{{\bf{t}}_2}} \right\}$   and $\left\{ {{{\bf{R}}_3},{{\bf{t}}_3}} \right\}$, we can have ${\bf{y}}_i^1 = {{\bf{R}}_1}{{\bf{x}}_i} + {{\bf{t}}_1}$, ${\bf{y}}_i^2 = {{\bf{R}}_2}{{\bf{x}}_i} + {{\bf{t}}_2}$ and ${\bf{y}}_i^3 = {{\bf{R}}_3}{{\bf{x}}_i} + {{\bf{t}}_3}$. Then we can find a plane ${\bf{\pi}}_i$ passing through ${\bf{y}}_i^1$, ${\bf{y}}_i^2$    and  ${\bf{y}}_i^3$, as demonstrated in Fig.~\ref{fig:ambiguous_l_pl} (b).  
	According to how we construct ${\pi _i}$, we know that $\left\{ {{{\bf{R}}_1},{{\bf{t}}_1}} \right\}$, $\left\{ {{{\bf{R}}_2},{{\bf{t}}_2}} \right\}$ and $\left\{ {{{\bf{R}}_3},{{\bf{t}}_3}} \right\}$  are three solutions for the  ${n_{\pi}}$ point-to-plane correspondences $\left\{ {{{\bf{x}}_i} \leftrightarrow {{\bf{\pi}}_i}} \right\}_{i = 1}^{{n_\pi}}$. 
	
\end{proof}

\begin{figure}	
	\centering	
	\includegraphics[width=0.48\textwidth]{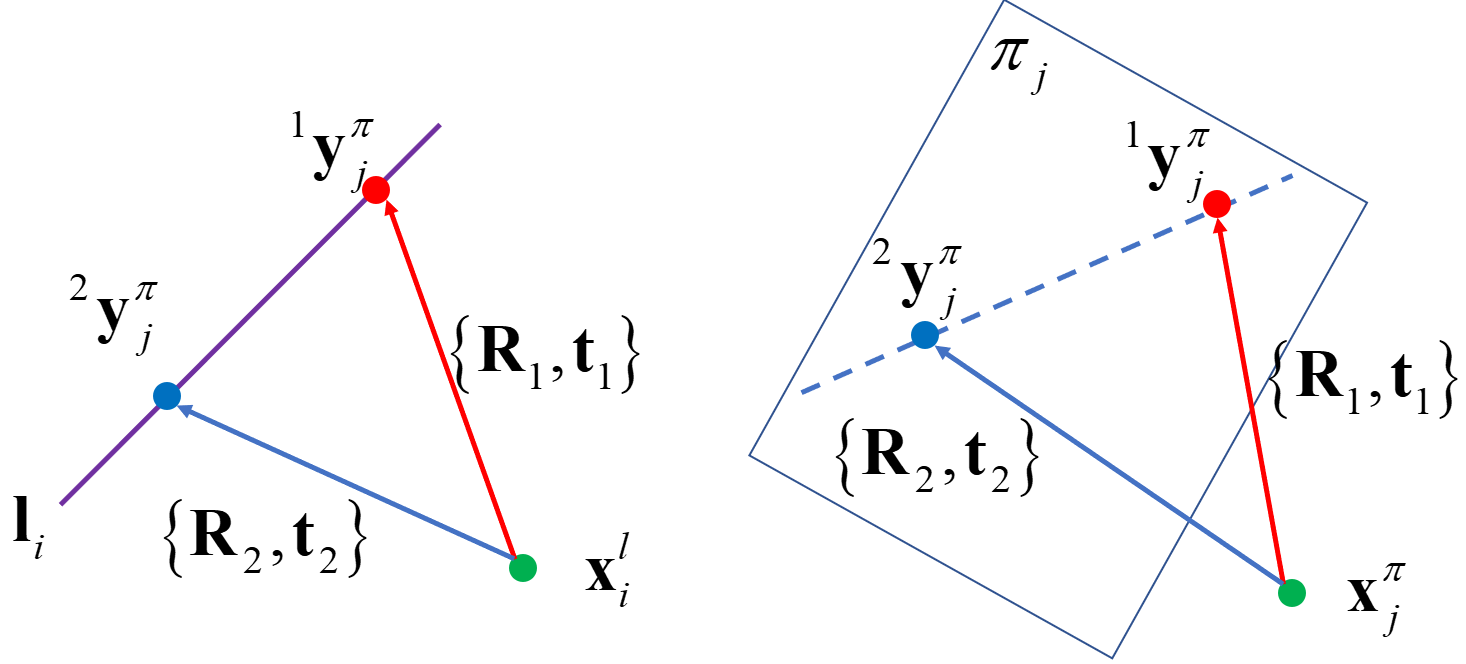}
	\caption{A schematic of point-to-line and point-to-plane ambiguous configuration.  $\left\{ {{{\bf{R}}_1},{{\bf{t}}_1}} \right\}$ and $\left\{ {{{\bf{R}}_2},{{\bf{t}}_2}} \right\}$ will transform ${\bf{x}}_i^l$ to ${\bf{l}}_i$,  and transform ${\bf{x}}_j^{\pi}$ to ${\bf{\pi}}_j$} 	\label{fig:ambiguous_l_and_pl}	
\end{figure}

	\noindent
	\textbf{Theorem 1.} For any $n_l$ points on lines, $n_{\pi}$ points on planes and any $\left\{ {{{\bf{R}}_1},{{\bf{t}}_1}} \right\}$ and $\left\{ {{{\bf{R}}_2},{{\bf{t}}_2}} \right\}$, there exist  $n_l$ lines and $n_{\pi}$ planes  to make $\left\{ {{{\bf{R}}_1},{{\bf{t}}_1}} \right\}$ and $\left\{ {{{\bf{R}}_2},{{\bf{t}}_2}} \right\}$ are exact solutions for the $n_l$ point-to-line and $n_{\pi}$ point-to-plane correspondences.

\begin{proof}
	We first consider the $n_l$ points on lines. Define ${\bf{x}}_i^{l}$ is the $i$th point. According to Lemma \ref{lemma:line}, we can find $n_l$ lines to make $\left\{ {{{\bf{R}}_1},{{\bf{t}}_1}} \right\}$ and $\left\{ {{{\bf{R}}_2},{{\bf{t}}_2}} \right\}$ are exact solutions for the $n_l$ point-to-line correspondences  $\left\{ {{{\bf{x}}_i^{l}} \leftrightarrow {{\bf{l}}_i}} \right\}_{i = 1}^{{n_l}}$.
	
	Then we consider the $n_{\pi}$ points on planes. For the $j$th point ${{\bf{x}}_j^{\pi}}$ within them, we can define ${}^1{\bf{y}}_j^\pi = {{\bf{R}}_1}{{\bf{x}}_j^{\pi}} + {{\bf{t}}_1} $ and ${}^2{\bf{y}}_j^\pi = {{\bf{R}}_2}{{\bf{x}}_j^{\pi}} + {{\bf{t}}_2} $. Let us denote ${\bf{\pi}}_j$ as a plane passing through the line defined by ${{}^1{\bf{y}}}_j^\pi$ and ${}^2{\bf{y}}_j^\pi$. According to how we construct ${\pi _j}$, we know that $\left\{ {{{\bf{R}}_1},{{\bf{t}}_1}} \right\}$ and $\left\{ {{{\bf{R}}_2},{{\bf{t}}_2}} \right\}$ are two solutions for the  ${n_\pi}$ point-to-plane correspondences $\left\{ {{{\bf{x}}_j^{\pi}} \leftrightarrow {{\bf{\pi}}_j}} \right\}_{j = 1}^{{n_\pi}}$.
	
	Therefore, $\left\{ {{{\bf{R}}_1},{{\bf{t}}_1}} \right\}$ and $\left\{ {{{\bf{R}}_2},{{\bf{t}}_2}} \right\}$ are the two solutions for the  $n_l$ point-to-line correspondences  $\left\{ {{{\bf{x}}_i^{l}} \leftrightarrow {{\bf{l}}_i}} \right\}_{i = 1}^{{n_l}}$ and    ${n_\pi}$ point-to-plane correspondences $\left\{ {{{\bf{x}}_i^{\pi}} \leftrightarrow {{\bf{\pi}}_i}} \right\}_{i = 1}^{{n_\pi}}$.
\end{proof}

According to Lemma~\ref{lemma:line} and Theorem~\ref{theorm:1}, our algorithm keeps the smallest 2 minimuma, if we  use  point-to-line correspondences or point-to-plane and point-to-line correspondences for pose estimation. If we only have point-to-plane correspondences, our algorithm keeps the smallest 3 minimuma , according to Lemma \ref{lemma:plane}.

\end{document}